\documentclass[journal]{IEEEtran}
\usepackage{amssymb}
\usepackage{cite}
\ifCLASSINFOpdf
\else
\fi
\usepackage{amsmath}
\usepackage{algorithmic}
\usepackage{array}
\usepackage{url}
\usepackage{amsthm}
\usepackage{booktabs}
\usepackage{graphicx}
\usepackage{subfigure}
\usepackage{epstopdf}
\usepackage{multirow}
\newtheorem{theorem}{Theorem}
\newtheorem{prop}[theorem]{Proposition}
\newtheorem{remark}[theorem]{Remark}

\begin{document}
\title{Generalized two-dimensional linear discriminant analysis with regularization}

\author{{Chun-Na~Li, Yuan-Hai~Shao,Wei-Jie~Chen, Zhen~Wang
        and~Nai-Yang~Deng}% <-this % stops a space
\thanks{C. N. Li is with the College of Zhijiang, Zhejiang University of Technology, Hangzhou, 310024, P.R.China. (e-mail: na1013na@163.com, lcn@zjc.zjut.edu.cn)}% <-this % stops a space
\thanks{Y. H. Shao is with the School of Economics and Management, Hainan University, Haikou, 570228, P.R.China. (e-mail: shaoyuanhai21@163.com. Corresponding author)}% <-this % stops a space
\thanks{W. J. Chen is with the College of Zhijiang, Zhejiang University of Technology, Hangzhou, 310024, P.R.China, and Centre for Artificial Intelligence, University of Technology Sydney, NSW, 2007, Australia. (e-mail: wjcper2008@126.com)}
\thanks{Z. Wang is with the School of Mathematical Sciences, Inner Mongolia University, Hohhot, 010021, P.R.China. (e-mail: wangz11@mails.jlu.edu.cn)}
%wangzhen@imu.edu.cn
\thanks{D. N. Yang is with the College of Science, China Agricultural University, Beijing, 100083, P.R.China (e-mail: dengnaiyang@cau.edu.cn)}}

% \thanks{Manuscript received April 19, 2005; revised August 26, 2015.}}


\markboth{Journal of \LaTeX\ Class Files,~Vol.~ , No.~ ,  ~ }%
{Shell \MakeLowercase{\textit{et al.}}: Generalized two-dimensional linear discriminant analysis with regularization}



\maketitle

\begin{abstract}
Recent advances show that two-dimensional linear discriminant analysis (2DLDA) is a successful matrix based dimensionality reduction method. However, 2DLDA may encounter the singularity issue theoretically, and also is sensitive to outliers. In this paper, a generalized Lp-norm 2DLDA framework with regularization for an arbitrary $p>0$ is proposed, named G2DLDA. There are mainly two contributions of G2DLDA: one is G2DLDA model uses an arbitrary Lp-norm to measure the between-class and within-class scatter, and hence a proper $p$ can be selected to achieve the robustness.
The other one is that the introduced regularization term makes G2DLDA enjoy better generalization performance and avoid the singularity.
In addition, an effective learning algorithm is designed for G2LDA, which can be solved through a series of convex problems with closed-form solutions.
Its convergence can be guaranteed theoretically when $1\leq p\leq2$. Preliminary experimental results on three contaminated human face databases show the effectiveness of the proposed G2DLDA.
\end{abstract}


% \fi
\IEEEpeerreviewmaketitle

\section{Introduction}

%1.第一段不提那么多了
%2. 提你的模型前说明与vector based 的方法相比，2D的这些问题都存在；但是研究不够充分
%3. 在第三四段的时候把2D的和1D的描述上再做些区分。跟第2点对应
\IEEEPARstart{D}{imensionality} reduction (DR) plays an important role in pattern recognition and has been studied extensively.
%There are several DR methods being widely used, such as principal component analysis (PCA) \cite{TurkPentland91}, locality preserving projection (LPP) \cite{LPP}, neighborhood preserving embedding (NPE) \cite{NPE}, and linear discriminant analysis (LDA) \cite{Fisher36, Fukunaga90, SwetsWeng96}.
For supervised DR, linear discriminant analysis (LDA) \cite{Fisher36, Fukunaga90} is usually employed to extract the most discriminative features. It finds the optimal discriminative direction by maximizing the between-class variance while simultaneously minimizing the within-class variance in the projected space.

However, when dealing with massive multi-dimensional data such as the real world two-dimensional (2D) face images, LDA often becomes inadequate due to the high-dimensionality and the loss of some useful natural structural information when converting multi-dimensional data to vector ones \cite{Kong05}.
Especially, when the data feature dimension is much larger than the number of samples, LDA may suffer from the small sample size (SSS) problem and hence encounters singularity.
To deal with these problems, matrix based LDA, i.e., 2DLDA \cite{Liuetal93,Kong05,LiYuan05,YangZhang05,XiongSwamy05,KongWangTeohetc05,Jingetal06} is studied.
Compared to LDA, 2DLDA alleviates the SSS problem when some mild condition is satisfied \cite{Kong05,LiYuan05}.
Even so, it may still encounter the singularity issue theoretically, which in turn will degenerate its performance. Moreover, both LDA and 2DLDA just maximize the between-class variance and minimize the within-class variance for the training data set, while do not consider the generalization ability on the test data. This over-fitting phenomenon arises from the fact that there is no control term on confidence interval for classical LDA and 2DLDA. Another problem existing in classical LDA and 2DLDA is that they are sensitive to the presence of outliers, because the L2-norm will exaggerate the effect of the data samples.

For the first issue, a popular method is the regularization technique, which replaces the within-class covariance matrix with a ridge-like covariance estimate, for example, the regularized LDA (RDA) \cite{Friedman89,Dudoit02,Bickel04,RLDA}. The regularization technique reduces the variance that associated with the sample based estimate, and hence stabilizes the estimate \cite{Friedman89}. In fact, it has been successfully applied when solving the ill-posed inverse problems \cite{Sullivan86}.
Meanwhile, the introduction of the regularization term controls the model complexity, and avoids over-fitting \cite{ChenYangetal13}.
Therefore, the regularization technique not only overcomes the singularity problem, but also leads better generalization ability.
%There are several variants of the regularized LDA \cite{Friedman89,Dudoit02,Bickel04}.
%The diagonal estimate with elements belonging to the within-class covariance matrix was considered in \cite{Friedman89,Dudoit02,Bickel04}.
%Guo et al. \cite{RLDA} proposed a RDA by adding an extra L2-norm regularization term to achieve robustness. The Guo's regularized LDA performs well in multivariate classification problems, and has been used widely.
%In practice, the regularization technique has been proved to be effective in many machine learning problems. For example, the Ridge regularization for least squares regression \cite{GolubHansen99}, the Lasso regularization \cite{Tibshirani96,WittenTibshirani11} and the elastic net for sparseness study \cite{ZouHastie05}, the Hessian regularization for image annotation \cite{LiuTao13}, the p-Laplacian regularization for sparse coding \cite{LiuZhaWang16}, the Schattern p-norm for recovering the low-rank \cite{RechtFazelParrilo10,XieGuLiu16}, the mixed regularization \cite{LiuTaoCheng14}.
For singularity and generalization problems in LDA, Lp-norm-like regularization with Ridge-like covariance estimate is a good choice \cite{Friedman89,Dudoit02,Bickel04,RLDA,Sullivan86,ChenYangetal13}.

For the second issue on the sensitivity to outliers, some approaches were also proposed. For vector based LDA, local Fisher discriminant analysis (LFDA) \cite{Sugiyama07},
%hat incorporates the spirit of locality preserving projections (LPP) into classical LDA to alleviate the effect of outliers by considering neighborhood information,
probability based minimax optimization technique \cite{LanckrietGhaoui03} and uncertainty LDA model \cite{KimMagnani05}. Since the application of L2-norm in LDA is one of the main reasons that causes LDA sensitive to outliers, the L1-norm based technique is also considered as an effective robust replacement,
%For example, L1-norm PCA and L1-norm two-dimensional PCA (2DPCA-L1)  \cite{Kwak08,LiPangYuan10,LietalPCA15,WangNie15},
%L1-norm DLPP and L1-norm two-dimensional DLPP \cite{Wangetal16LPP,LuYuanLPP17} are proved to be more robust than the corresponding L2-norm based methods.
%In particular, the L1-norm methods have also shown their robustness on LDA and 2DLDA,
such as the rotational invariant L1-norm based LDA (DCL1) \cite{LDAR1} and the L1-norm based LDA (LDA-L1) \cite{ZhongZhang13,WangTZ12}.
Here we notice that the solving algorithm of LDA-L1 was based on the gradient ascending (GA) technique of nonconvex surrogate functions whose optimal solutions cannot be guaranteed, and the proper step size was hard to choose in practice.
To tackle this problem, various methods were proposed, including the non-greedy iterative algorithms for difference optimization problems \cite{ChenYangetal14,YeYangLiu16,Liu17}, the convex surrogate technique \cite{Zheng14}, the concave-convex procedure (CCCP) \cite{YeZhaoetc12} and the successive linear algorithm (SLA) \cite{AVIDA}. Though the above improvements were proved to be effective, it should be noted that some of them still exist the singularity problem during practical computation, for example, L1-LDA \cite{Zheng14} and recursive ``concave-convex" Fisher linear discriminant (RPFLD) \cite{YeZhaoetc12}, as pointed out in \cite{ChenYangetal13} and \cite{YeYangLiu16}, respectively.
Further, the generalization of L1-norm LDA to Lp-norm LDA with any $p>0$ was studied \cite{KwakLpLDA,An14}, and the Lp-norm LDA was solved through the GA technique. For 2DLDA, as a robust improvement of LDA, LDA-L1 was further extended to its 2D version named L12DLDA \cite{L12DLDA15,ChenL12DLDA15}, and its non-greedy modification \cite{TrL12DLDA17} was also studied.

However, for the singularity problem of 2DLDA and its generalization and robustness issues, they are not well addressed as LDA. In particular, the regularization technique and the application of Lp-norm are not studied as far as we know. Therefore,
in this paper, to address the singularity problem of 2DLDA and improve its generalization and robustness performance, we consider a generalized Lp-norm based 2DLDA with regularization for an arbitrary $p>0$, named G2DLDA.
G2DLDA not only maximizes the Lp-norm between-class distance and minimizes the Lp-norm within-class distance, but also introduces a regularization term with Lp-norm.
Since the proposed G2DLDA involves the Lp-norm operation on both its numerator and denominator, we here employ a simple iteration technique which converts the ratio optimization problem into a series of convex programming problems, and its convergence is also discussed. In summary, the proposed G2DLDA has following characteristics:\\
(i) G2DLDA is a generalized two-dimensional linear discriminant analysis with regularization, where the between-class scatter, within-class scatter and the regularization term in G2DLDA are measured by Lp-norm with arbitrary $p>0$. This makes G2DLDA not only easily degenerate to the existing 2DLDA and L12DLDA, but also achieve desired performance by choosing proper $p$.\\
(ii) The regularization is used to remedy the singularity problem. In addition, it controls the model complexity and avoids over-fitting, and therefore improves its generalization performance.\\
(iii) An effective algorithm is designed for G2DLDA. In specific, the solution of G2DLDA  is given by solving a series of convex problems with closed-form solutions. Moreover, the convergence of the algorithm can be ensured when $1\le p\le2$.\\
(iv) Experimental results on three contaminated human face databases  with different noise levels demonstrate the effectiveness of G2DLDA.

%(i) The between-class scatter and the within-class scatter in G2DLDA are measured by arbitrary norm, and hence different norms can be chosen to achieve desired performance.
%
%(ii) The G2DLDA criterion introduces the regulation technique, which is the first time for 2DLDA related methods. The regularization term makes G2DLDA avoid the singularity problem. In addition, it controls the model complexity and avoids over-fitting, and therefore improves its generalization performance.
%
%(iii) The norm used for the regularization term is not constraint to the L2-norm as usual. In fact, the Lp-norm as in the between-class and within-class scatters is employed. Therefore, we can select the appropriate regularization term by varying $p$.
%
%(iv) G2DLDA is a general framework, and can be degenerated to 2DLDA and L12DLDA.
%
%(v) Though the G2DLDA formulation is of the Lp-norm ratio form for arbitrary $p$, it can be solved through a series of convex problem, with analytic solution is given in each iteration. Moreover, the convergence of the algorithm can be ensured when $1\leq p\leq2$.
%
%(vi) Experimental results on three contaminated human face databases demonstrate the effectiveness of G2DLDA.

The paper is organized as follows. Section 2 briefly dwells on the LDA and 2DLDA. Section 3 reviews L1LDA and L12DLDA.
Section 4 proposes our G2DLDA and gives the corresponding theoretical analysis. Section 5 makes comparisons of our G2DLDA with its related approaches. At last, the concluding remarks are given in Section 6.

The notations of this paper are given as follows. We consider a supervised learning problem in the $(d_1\times d_2)$-dimensional matrix space $\mathbb{R}^{d_1\times d_2}$. The training data set is given by $T=\{(X_{1},y_{1}),...,(X_{N},y_{N})\}$, where $X_{l}\in\mathbb{R}^{d_1\times d_2}$ is the input matrix and $y_{l}\in \{1,...,c\}$ is the corresponding label with $l=1,...,N$. Assume that the $i$-th class contains $N_i$ samples, $i=1,\ldots,c$. Then we have $\sum\limits_{i=1}^{c}N_i=N$. We further write the samples in the $i$-th class as $\{X_{ij}\}$, $i=1,\ldots,c$, $j=1,\ldots,N_i$. Let $\overline{X}=\frac{1}{N}\sum\limits_{l=1}^{N}X_l$ be the mean of all sample matrices and ${\overline{X}}_i=\frac{1}{N_i}\sum\limits_{j=1}^{N_i}X_{ij}$ be the mean of sample matrices in the $i$-th class. For a matrix $Q=(q_1,\,q_2,\ldots,q_n)\in R^{m\times n}$ and $p>0$, its Lp-norm $||Q||_p$ is defined as $||Q||_p=(\sum\limits_{i=1}^{n}||q_i||_p^p)^{\frac{1}{p}}$. Note that when $0<p<1$, Lp-norm is only a quasi-norm \cite{Pekalska06}. However, it does not affect its use in this paper, and we hence call it Lp-norm for symbol unification.

\section{L2-norm LDA and L2-norm 2DLDA}\label{2DLDA}
%We further organize the training samples in $T$ as $X=(x_1,\,x_2,\ldots,x_N)\in\mathbb{R}^{n\times N}$
%We organize the training points as $X\in\mathbb{R}^{m\times n}$.
The classical L2-norm based LDA is arising from Fisher's discriminant problem, and is a vector based method. Assume $d_2=1$, then each $X_i$ lies in the $d_1$-th dimensional vector space $\mathbb{R}^{d_1}$. For the data set $T$, define the between-class scatter matrix and the within-class scatter matrix as
\begin{equation}\label{Sb}
S_b=\frac{1}{N}\sum\limits_{i=1}^{c}N_i({\overline{X}}_i-{\overline{X}})({\overline{X}}_i-{\overline{X}})^T
\end{equation}
and
\begin{equation}\label{Sw} S_w=\frac{1}{N}\sum\limits_{i=1}^{c}\sum\limits_{j=1}^{N_i}(X_{ij}-{\overline{X}}_i)(X_{ij}-{\overline{X}}_i)^T.
\end{equation}

LDA involves seeking an optimal matrix $W=[w_1,\,w_2,\ldots,w_{r_1}]\in\mathbb{R}^{d_1\times r_1}$ that consists of
discriminant vectors $w_i\in\mathbb{R}^{d_1}$, $i=1,2,\ldots,r_1$, $r_1\leq d_1$, by solving the problem
\begin{equation}\label{LDA}
\begin{split}
\underset{W}{\max}&~~\frac{\hbox{tr}(W^TS_bW)}{\hbox{tr}(W^TS_wW)}.
\end{split}
\end{equation}

After obtaining $W$, a new coming input $X$ is projected as $C=W^TX$. The solution to the optimization problem \eqref{LDA} can be given by the eigenvectors corresponding to the first largest $s$ nonzero eigenvalues of the generalized eigenvalue problem $S_bw=\lambda S_ww$ in case $S_w$ is nonsingular, where $\lambda\not=0$. Since the rank of $S_b$ is at most $c-1$, the number of extracted features is less or equal than $c-1$. It is obvious that when $S_w$ is not of full rank, LDA will encounter the singularity problem.

To deal with the matrix data directly, 2DLDA is studied. Assume $d_2>1$ and the input data are as demonstrated in $T$. Then 2DLDA has the same formulation of LDA \eqref{LDA}, where $S_b$ and $S_w$ are defined as in \eqref{Sb} and \eqref{Sw} but with the matrix input. The projection vectors of 2DLDA are also obtained by solving the above generalized eigenvalue problem.
2DLDA can well extract algebraic features of the matrix input data \cite{Liuetal93}. Though the singularity problem is much alleviated for 2DLDA, it still exists in theory \cite{LiYuan05}.

In addition, both LDA and 2DLDA are prone to the presence of outliers. In fact, denote $V_i={\overline{X}}_i-{\overline{X}}\in\mathbb{R}^{d_1\times d_2},~~Z_{ij}=X_{ij}-{\overline{X}_i}\in\mathbb{R}^{d_1\times d_2},$
$i=1,\ldots,c$, $j=1,\ldots,N_i$, where $X_{ij},\,{\overline{X}_i},\,{\overline{X}}$ are defined as in Section 1.
Then by the fact that $\hbox{tr}(S S^T)=\|S\|_F^2$ for any matrix $S$, where $\|\cdot\|_F$ is the Frobenius norm, \eqref{LDA} can be rewritten as
\begin{equation}\label{2DLDAL2Fnorm}
\begin{split}
\underset{W}{\max}~~&\frac{\sum\limits_{i=1}^{c}N_i\|W^TV_{i}\|_F^2}{\sum\limits_{i=1}^{c}\sum\limits_{j=1}^{N_i}\|W^TZ_{ij}\|_F^2}.
\end{split}
\end{equation}
%Since for a matrix $Q=(q_1,\,q_2,\ldots,q_n)\in R^{m\times n}$, its F-norm is defined as $||Q||_F=\sqrt{\sum\limits_{i=1}^{n}||q_i||_2^2}$.
As can be seen, the objective of \eqref{2DLDAL2Fnorm} is based on the L2-norm in nature, and hence is sensitive to outliers and noise.

%Note that the definition of $S_b$ is based on the $L_2$-norm, and therefore outliers will affect the between-class distance, hence it is sensitive to outliers. Besides, since LDA solves generalized eigenvalue problem, it may encounter the SSS problem.

\section{L1-norm LDA and L1-norm 2DLDA}\label{L12DLDA}
%In order to reduce the sensitivity, an L$_1$-norm based 2DLDA is proposed by replacing the L$_2$-norm by the L$_1$-norm \cite{L12DLDA15}, which is formulated as follows
With the purpose to improve the robustness of LDA and 2DLDA, L1-norm based LDA and 2DLDA (LDA-L1 and L12DLDA) are formulated by replacing the F-norm terms in \eqref{2DLDAL2Fnorm} by the L1-norm ones:

\begin{equation}\label{2DLDAL1W}
\begin{split}
\underset{W}{\max}~~&\frac{\sum\limits_{i=1}^{c}N_i\|W^TV_{i}\|_1}{\sum\limits_{i=1}^{c}\sum\limits_{j=1}^{N_i}\|W^TZ_{ij}\|_1}\\
\hbox{s.t.}~~&W^TW=I,
\end{split}
\end{equation}
where $W=[w_1,\ldots,w_{r_1}]\in\mathbb{R}^{d_1\times r_1}$ is the orthonormal projection matrix.
When $d_2=1$, \eqref{2DLDAL1W} is LDA-L1 \cite{ZhongZhang13,WangTZ12}. For general $d_2>1$, it is just L12DLDA \cite{L12DLDA15,ChenL12DLDA15}. When $r_1=1$, \eqref{2DLDAL1W} is originally solved through the GA technique for a nonconvex surrogate function, and the deflation technique is used to obtain multiple discriminant directions. Note that GA needs to choose an appropriate step size, and the optimal solution of \eqref{2DLDAL1W} can not be guaranteed in GA technique \cite{Zheng14}.

To address this problem, when $d_2=1$, \cite{Zheng14} derived a novel L1-norm discriminant criterion coined L1-LDA under a theoretical framework of Bayes optimality. L1-LDA is of the same formulation of LDA-L1 but uses the whole data scatter $S_t$ instead of the within-class scatter, all $N_i$ are equal, and the orthonormal constraint in \eqref{2DLDAL1W} is replaced by the $S_t$-orthonormal one. L1-LDA is solved by an iteration technique, and the $(t+1)$-th solution $w^{(t+1)}$ is given by \begin{equation}\label{ZhengL1LDA2}
\begin{split}
w^{(t+1)}&=\arg\underset{w}{\min}~~w^TV_t(w^{(t)})w\\
\hbox{s.t.}~~&\sum\limits_{i=1}^{c}s_i^{(t)}w^TB_i=1,\\
\end{split}
\end{equation}
where $V_t(w^{(t)})=\sum\limits_{l=1}^{N}M_lM_l^T/|(w^{(t)})^TM_l|$, $M_l=X_{l}-{\overline{X}}$, $B_i={\overline{X}}_{i}-{\overline{X}}$, and $s_i^{(t)}=\textrm{sign}((w^{(t)})^TB_i)$.
Here $\textrm{sign}(\cdot)$ is the symbol function.

%Define
%$M_{l}=X_{l}-{\overline{X}}\in\mathbb{R}^{d_1}$, $l=1,2,\ldots,N$ and $M=[M_1,M_2,\ldots,M_{N}]\in\mathbb{R}^{d_1\times N}$. Then the $s$-th discriminant vector is defined as the solution of the following problem:
%\begin{equation}\label{ZhengL1LDA}
%\begin{split}
%\underset{w}{\max}~~&\frac{\sum\limits_{i=1}^{c}N_i\|w^TV_{i}\|_1}{\sum\limits_{l=1}^{N}\|w^TM_{l}\|_1}\\
%\hbox{s.t.}~~&w^TMM^Tw_j=0,\\
%&j=1,2,\ldots,s-1.
%\end{split}
%\end{equation}
%When $s=1$, problem \eqref{ZhengL1LDA} is solved through a series of quadratic problems.
%%When $s=1$, problem \eqref{ZhengL1LDA} is rewritten as
%%\begin{equation}\label{ZhengL1LDA2}
%%\begin{split}
%%\underset{w}{\min}~~&{\sum\limits_{l=1}^{N}\|w^TM_{l}\|_1}\\
%%\hbox{s.t.}~~&{\sum\limits_{i=1}^{c}N_i\|w^TV_{i}\|_1}=1,
%%\end{split}
%%\end{equation}
%%and is solved through a serious of quadratic problems.
%Specifically, suppose $w^{(t)}$ is the optimal projection vector in the $t$-th iteration. Then $w^{(t+1)}$ is given by
%\begin{equation}\label{ZhengL1LDA2}
%\begin{split}
%w^{(t+1)}&=\arg\underset{w}{\min}~~w^TV_t(w^{(t)})w\\
%\hbox{s.t.}~~&\sum\limits_{i=1}^{c}s_i^{(t)}w^TV_i=1,\\
%\end{split}
%\end{equation}
%where $V_t(w^{(t)})=\sum\limits_{l=1}^{N}M_lM_l^T/|(w^{(t)})^TM_l|$, $M_l$ is the $l$-th column of $M$, and $s_i^{(t)}=\textrm{sign}((w^{(t)})^TV_i)$.
%Here $\textrm{sign}(\cdot)$ is the symbol function.
%%defined as
%%\begin{equation*}\label{}
%%\begin{split}
%%\textrm{sign}(a)=\begin{cases}1,~~~~a>0,\\0,~~~~a=0,\\-1,~~~~a<0.\end{cases}
%%\end{split}
%%\end{equation*}

Let $B=[B_1,B_2,\ldots,B_c]$ and $s^{(t)}=[s_1^{(t)},s_2^{(t)},\ldots,s_c^{(t)}]^T$. Then the solution of \eqref{ZhengL1LDA2} is given by
\begin{equation}\label{ZhengL1LDA4}
\begin{split}
w^{(t+1)}&=\frac{[V_t(w^{(t)})]^{-1}Bs^{(t)}}{(Bs^{(t)})^T[V_t(w^{(t)})]^{-1}(Bs^{(t)})}.
\end{split}
\end{equation}

Though L1-LDA is derived under the rigorous theoretical framework, it is obvious that the matrix $V_t(w^{(t)})$ in the solution \eqref{ZhengL1LDA4} of each iteration may not be of full rank, and hence $(V_t(w^{(t)}))^{-1}$ may not exist. The singularity problem happens because the matrix in the quadratic objective of problem \eqref{ZhengL1LDA2} may not be positive definite. This will bring the ineffectiveness of the algorithm. The phenomenon is also theoretically stated in \cite{YeYangLiu16}.
Moreover, it can be seen that problem \eqref{2DLDAL1W} just minimizes the empirical error for the training data. However, it does not consider the generalization property.
As L12DLDA is also solved by GA, L12DLDA exists the same problem as LDA-L1. However, there is no study on this problem on L12DLDA. If we try to address it as did in \cite{Zheng14}, L12DLDA can also be solved through a series of convex quadratic problems. However, it will also encounter the singularity problem. Thereafter, it is necessary to study the above problems further.
%Note that problem \eqref{ZhengL1LDA3} is transformed from problem \eqref{ZhengL1LDA2}, therefore, we examine problem \eqref{ZhengL1LDA2}.

%After obtaining $W$, a new coming input $X$ is projected as $C=W^TX$.

%In fact, the projective transformation can also be considered as
%$C_1=XV$ as did in \eqref{2DLDAL2Fnorm}, and the problem is given as
%\begin{equation}\label{2DLDAL1V}
%\begin{split}
%\underset{W}{\max}~~&\frac{\sum\limits_{i=1}^{c}N_i\|W^TY_{i}\|_1}{\sum\limits_{i=1}^{c}\sum\limits_{j=1}^{N_i}\|W^TZ_{ij}\|_1}\\
%\hbox{s.t.}~~&W^TW=I,
%\end{split}
%\end{equation}
%where $W=[w_1,\,w_2,\ldots,w_{r_1}]\in\mathbb{R}^{d_1\times r_1}$.
%In the following, when we refer L12DLDA, we mean problem \eqref{2DLDAL1V}.

\section{Generalized Lp-norm 2DLDA with regularization}
\subsection{Problem formulation}

As seen in Sections \ref{2DLDA} and \ref{L12DLDA}, the above L2-norm and L1-norm based ratio form in LDAs and 2DLDAs face the singularity problem. Moreover, though they may perform well for the training data set, their optimization models did not consider the generalization ability. This phenomenon arises from the fact that there is no regularization term on classical LDA and 2DLDA. Aiming to solve the above singularity problem, and improve the robustness and generalization performance of 2DLDA, we here propose a generalized Lp-norm based 2DLDA for arbitrary $p>0$ with regularization, called G2DLDA, which is formulated as
%with the purpose to suppress the influence of outliers and noise more effectively.

\begin{equation}\label{G2DLDAV0}
\begin{split}
\underset{W}{\min}~~&\frac{\sum\limits_{i=1}^{c}\sum\limits_{j=1}^{N_i}\|W^TZ_{ij}\|_p^p+\sigma||W||_p^p}{\sum\limits_{i=1}^{c}N_i\|W^TV_{i}\|_p^p}\\
\hbox{s.t.}~~&W^TW=I,\\
\end{split}
\end{equation}
where $W\in R^{d_1\times r_1}$, and $\sigma$ is a nonnegative tuning parameter.

We now explain the geometric meaning of problem \eqref{G2DLDAV0}.
By minimizing the first term in the nominator, each element in the $i$-th class is guaranteed to be as close as possible to the $i$-th class center, while maximizing the denominator makes each projected class center as far as possible from the whole projected center in the Lp-norm sense. Minimizing the second term in the nominator, leads us to control the model complexity, which will generate better generalization ability.
The constraint of \eqref{G2DLDAV0} makes sure the obtained discriminant directions of G2DLDA orthogonal to each other, which is beneficial to the nonredundancy in representing the subspace.

As we can see from problem \eqref{G2DLDAV0}, there are two improvements of G2DLDA over the existing two-dimensional LDAs: (i) introducing an arbitrary Lp-norm to measure the between-class scatter and the within-class scatter and (ii) minimizing an extra regularization term in 2DLDA.

\begin{remark}
Problem \eqref{G2DLDAV0} can be viewed as a general framework for two-dimensional LDA. By choosing different $p$ and $\sigma$, we obtain different existing two-dimensional LDA models. In particular,
%(i) when $\sigma=0$, $p=2$, and $d_2=1$, then G2DLDA is LDA.
%
%(i) when $\sigma=0$, $p=1$, and $d_2=1$, then G2DLDA is LDA-L1.
%
%(iii) when $\sigma>0$, $p=2$, and $d_2=1$, then G2DLDA is RDA.
when $\sigma=0$, $p=2$, and $d_2>1$, then G2DLDA is 2DLDA;
when $\sigma=0$, $p=1$, and $d_2>1$, then G2DLDA is L12DLDA.
Further, when $d_2=1$, G2DLDA degenerates to the vector based LDA. In this situation, when $\sigma=0$ and $p=2$, G2DLDA is LDA; when $\sigma=0$ and $p=1$, G2DLDA is LDA-L1; when $\sigma=0$ and $p>0$, G2DLDA is LDA-Lp. In particular, when $d_2=1$, $\sigma>0$ and $p=2$, G2DLDA is just RDA, which is an important improvement of LDA by considering an extra regularization term.
In RDA, the regularization term not only makes it avoid the SSS problem, %where the sample covariance matrix estimate may be singular and hence not invertible,
but also makes direct matrix operation available for high dimensionality \cite{RLDA}. Moreover, it helps controlling the model complexity and hence avoids the over-fitting problem \cite{ChenYangetal13}.
\end{remark}

However, for matrix based LDA, there is no corresponding regularized model.
To give a clearer picture of the regularization term in G2DLDA, we reformulate problem \eqref{G2DLDAV0} as:
\begin{equation}\label{G2DLDAV}
\begin{split}
\underset{W}{\min}~~&\sum\limits_{i=1}^{c}\sum\limits_{j=1}^{N_i}\|W^TZ_{ij}\|_p^p+\sigma||W||_p^p\\
\hbox{s.t.}~~&{\sum\limits_{i=1}^{c}N_i\|W^TV_{i}\|_p^p}=1,\\
~~&W^TW=I.\\
\end{split}
\end{equation}
By observing problem \eqref{G2DLDAV}, we see G2DLDA can be separated into two parts: the first part includes the first term in the objective together with the constraint, which realizes minimizing the empirical error on the training data, while the second part is the regularization term in the objective, which makes sure our G2DLDA not only works on the training data, but also controls the confidence interval of our model.
%considers the generalization property on the test data.
%For the first part, by using an appropriate $p$ for different data sets, G2DLDA will possess fair robustness.
%For the second part,
The regularization technique is in fact applied extensively in many pattern recognition methods, such as some great improvements on support vector machines (SVMs) \cite{TBSVM,LSPTSVM} with L2-norm-like regularization terms, which realized the structural risk minimization.
%For example, by adding regularization terms, the twin bounded support vector machines (TBSVM) \cite{TBSVM} and the least squares projection twin support vector machine (LSPTSVM) \cite{LSPTSVM} improved the twin support vector machine (TWSVM) \cite{TWSVM} and the projection twin support vector machine (PTSVM) \cite{PTSVM}, respectively, ensuring the optimization problems in TBSVM and LSPTSVM are positive definite, which was benefit for deriving their dual problems.
%Moreover, the regularization terms realized the minimization of structural risk, where TWSVM and PTSVM only considered the empirical risk.
%Further, Ye \cite{Ye07} presented the equivalence relationship between the multivariate linear regression and RDA, which was shown empirically successfully applicable in many applications involving high-dimensional data.
For our G2DLDA, the regularization term on one hand can control the model complexity and hence generate better generalization ability. On the other hand, we will see that in the following solving procedure of G2DLDA, this additional regularization term conquers the singularity problem that exists in \eqref{ZhengL1LDA2}
%, which is similar to that of \cite{TBSVM,LSPTSVM,Ye07}.
The influence of different $p$ and the effect of the regularization term to our G2DLDA will also be investigated experimentally.
\subsection{The solving of the proposed G2DLDA for one projection}
The problem \eqref{G2DLDAV0} is in the ratio form, and both its denominator and nominator contain the Lp-norm. Therefore, it is hard to obtain all the projection directions once for all. Hence, we first consider the corresponding problem with one projection vector
\begin{equation}\label{G2DLDAv0}
\underset{w}{\min}~~J_0(w)=\frac{\sum\limits_{i=1}^{c}\sum\limits_{j=1}^{N_i}\|w^TZ_{ij}\|_p^p+\sigma||w||_p^p}{\sum\limits_{i=1}^{c}N_i\|w^TV_{i}\|_p^p},
\end{equation}
subject to $w^Tw=1$, where $w\in\mathbb{R}^{d_1}$.

As in \eqref{G2DLDAV}, we rewrite problem \eqref{G2DLDAv0} as
\begin{equation}\label{}
\begin{split}
\underset{w}{\min}~~&{\sum\limits_{i=1}^{c}\sum\limits_{j=1}^{N_i}\sum\limits_{k=1}^{{d_2}}|w^TZ_{ijk}|^p+\sigma||w||_p^p}\\
\hbox{s.t.}~~&{\sum\limits_{i=1}^{c}\sum\limits_{k=1}^{{d_2}}N_i|w^TV_{ik}|^p}=1.
\end{split}
\end{equation}
The above problem is equivalent to
\begin{equation}\label{appromodelRp}
\begin{split}
\underset{w}{\min}~~&w^T\big(\sum\limits_{i=1}^{c}\sum\limits_{j=1}^{N_i}\sum\limits_{k=1}^{{d_2}}\frac{Z_{ijk} Z_{ijk}^T}{|w^TZ_{ijk}|^{2-p}}+\sigma\cdot{diag\big(\frac{1}{|w|_k^{2-p}}\big)}\big)w\\
\hbox{s.t.}~~&{\sum\limits_{i=1}^{c}\sum\limits_{k=1}^{{d_2}}N_i|w^TV_{ik}|^{p-1}\textrm{sign}(w^TV_{ik})w^TV_{ik}}=1,
\end{split}
\end{equation}
where ${diag\big(\frac{1}{|w|_k^{2-p}}\big)}$ represents the diagonal matrix with its $(k,k)$-th element $\frac{1}{|w|_k^{2-p}}$, $k=1,2,\ldots,d_1$.

We now present an iterative algorithm to solve \eqref{appromodelRp}. Let $t$ be the iteration number. Denote
\begin{equation}\label{HtRp}
H^{(t)} = \sum\limits_{i=1}^{c}\sum\limits_{j=1}^{N_i}\sum\limits_{k=1}^{d_2}\frac{Z_{ijk} Z_{ijk}^T}{|({w^{(t)}})^TZ_{ijk}|^{2-p}}+{\sigma\cdot diag\big(\frac{1}{|{w^{(t)}}|_k^{2-p}}\big)},
\end{equation}
\begin{equation}\label{htRp}
h^{(t)} = {\sum\limits_{i=1}^{c}\sum\limits_{k=1}^{d_2}N_i|({w^{(t)}})^TV_{ik}|^{p-1}\textrm{sign}(({w^{(t)}})^TV_{ik})V_{ik}}.
\end{equation}
Then we solve the following problem to get $w^{(t+1)}$:
\begin{equation}\label{appromode2Rp}
\begin{split}
\underset{w}{\min}~~&w^TH^{(t)}w\\
\hbox{s.t.}~~&({h^{(t)}})^Tw=1.
\end{split}
\end{equation}
By the definition of $H^{(t)}$, we see that the regularization term makes it positive definite. Note that to make sure $H^{(t)}$ is well defined, it requires that $|w^TZ_{ijk}^{(t)}|\not=0$ and $|w^{(t)}|_k\not=0$. If this happens, we should let $w^{(t)}=(w^{(t)}+\delta)/||w^{(t)}+\delta||_2^2$, where $\delta$ is a small random vector.

Now we solve problem \eqref{appromode2Rp}. Let the Lagrangian of \eqref{appromode2Rp} be
\begin{equation}\label{LagRp}
L(w,~ \lambda)=w^TH^{(t)}w-\lambda((h^{(t)})^Tw-1).
\end{equation}
Then the corresponding KKT conditions are
\begin{equation}\label{KKTRp}
2H^{(t)}w-\lambda h^{(t)}=0,~~(h^{(t)})^Tw-1=0.
\end{equation}
From \eqref{KKTRp}, we get $\lambda=2((h^{(t)})^T (H^{(t)})^{-1}h^{(t)})^{-1}$.
%Then the KKT condition is
%\begin{equation}\label{KKT1Rp}
%2H^{(t)}w-\lambda h^{(t)}=0,
%\end{equation}
%\begin{equation}\label{KKT2Rp}
%(h^{(t)})^Tw-1=0.
%\end{equation}
%From \eqref{KKT1Rp}, we get $w=\frac{1}{2}\lambda (H^{(t)})^{-1}h^{(t)}$. By combining \eqref{KKT2Rp}, we have $\frac{1}{2}\lambda(h^{(t)})^T (H^{(t)})^{-1}h^{(t)}=1$, which gives $\lambda=2((h^{(t)})^T (H^{(t)})^{-1}h^{(t)})^{-1}$.
Therefore,
\begin{equation}\label{wtRp}
w=\frac{(H^{(t)})^{-1}h^{(t)}}{(h^{(t)})^T (H^{(t)})^{-1}h^{(t)} }.
\end{equation}
Now we summarize the above procedure in Algorithm 1.
\begin{center}
\begin{tabular}{lllll}
\toprule
\noindent{$\mathbf{Algorithm~1.}$~ G2DLDA solving algorithm for one }\\
\quad\quad\quad\quad\quad\quad\quad  {direction }\\
\midrule
\textbf{Input}: The training data set $T=\{(X_{1},y_{1}),...,(X_{N},y_{N})\}$ \\with $X_l \in R^{{d_1}\times d_2}$, $l=1,2,\ldots, N$, parameter $\sigma>0$, \\ stopping criterion $\epsilon>0$ and maximum iteration number \\ $itmax$.\\

I. \textbf{Initialization}. Set the iteration number $t=0$ and  \\
\quad $w^{(0)}=\frac{\textbf{1}}{||\textbf{1}||_2}$,
where $\textbf{1}$ is the vector of all ones.\\
II. \textbf{Repeat}\\

\quad (a) Compute $H^{(t)}$ and $h^{(t)}$ according to \eqref{HtRp} and \eqref{htRp}, \\
\quad\quad\quad respectively.\\

\quad (b) Compute $w^{(t+1)}$ according to \eqref{wtRp}, and let\\
\quad \quad \quad $w^{(t+1)}=\frac{w^{(t+1)}}{||w^{(t+1)}||_2^2}$.\\

\quad (c) Set $t=t+1$.\\
\textbf{Until} $||w^{(t+1)}-w^{(t)}||_2<\epsilon$ or $t>itmax$.\\
\noindent {\bf Output:} Discriminant vector $w^*=w^{(t+1)}$.\\
\bottomrule
\end{tabular} \label{Algorithm1R}\\
\end{center}
Next, we show that the above Algorithm 1 is convergent when $1\leq p\leq 2$.
\begin{prop}\label{convergeth}
When $1\leq p\leq 2$, Algorithm 1 monotonically decreases the objective of
problem \eqref{G2DLDAv0} in each iteration, and hence converges.
\end{prop}
\noindent\textbf{Proof:}
The proof is in the supplemenary material. \hfill $\square$

We now analyze the time complexity of 2DLDA, L12DLDA and the proposed algorithm when one discriminant discriminant is learnt. Denote $N$ the number of samples, the number of classes $c$, the feature space $d_1\times d_2$, $d=\max(d_1,d_2)$, and $T$ the iteration number.
%For 2DPCA \cite{YangZhang04} and L12DPCA \cite{LiPangYuan10}, their time complexity is $O(d^3)$ and $O(T(Nd_1d_2))$, respectively.
For 2DLDA \cite{Liuetal93}, the time complexity is $O((N+c+1)d)$. The time complexity of L12DLDA \cite{L12DLDA15,ChenL12DLDA15} is $O(T((N+c) d_1d_2))$. For our G2DLDA, the time complexity to compute $H(t)$ and $h(t)$ in (13) and (14) is $O(cNd_1d_2)$ and $O(cd_1d_2)$. To obtain $w$ in (18), it needs $O(d^3)$. Therefore, the total complexity of G2DLDA is $O(T(cNd_1d_2 +d^3))$.

\subsection{G2DLDA for multiple orthogonal projection directions }
By performing Algorithm 1, we get one projection vector.
If we want to project the data into multi-dimensional space, more orthogonal projection vectors are needed.
%Algorithm 2 will finish the job.
Suppose we have obtained the first orthogonal $s$ discriminant vectors $w_1,\,w_2,\ldots,w_{s}$, $w_i\in\mathbb{R}^{d_1}$. To compute the next projection vector $w_{s+1}$ by minimizing $J_0(w)$, it needs to satisfy the following orthogonal constraints \cite{LFDP}
\begin{equation}\label{}
w_1^Tw_{s+1}=w_2^Tw_{s+1}=\cdots=w_s^Tw_{s+1}=0.
\end{equation}

Denote $W_s=[w_1,\,w_2,\ldots,w_{s}]\in\mathbb{R}^{d_1\times s}$ and let $U_s=span \{w_1,\,w_2,\ldots,w_{s}\}$ be the linear subspace with $dim(U_s)=s$. It is obvious that $w_{s+1}$ is required to be orthogonal to $U_s$. Give a basis matrix $B=[b_1,\,b_2,\ldots,b_{d_1-s}]\in\mathbb{R}^{d_1\times (d_1-s)}$ of the space $U_s^{\bot}$, where $U_s^{\bot}$ is the null space of $U_s$ with $dim(U_s^{\bot})=d_1-s$.
Then, we can solve the following problem
\begin{equation}\label{G2DLDAvorth0}
\underset{w\in U_s^{\bot}}{\min}~~\frac{\sum\limits_{i=1}^{c}\sum\limits_{j=1}^{N_i}\|w^TB^TZ_{ij}\|_p^p+\sigma||w||_p^p}{\sum\limits_{i=1}^{c}N_i\|w^TB^TV_{i}\|_p^p}
\end{equation}
to obtain $w_{s+1}$. Note that the above problem projects the data onto $U_s^{\bot}$, and therefore the computation of $w_{s+1}$ is in a $(d_1-s)$-dimensional space. This leads the output to be orthogonal to any vectors in $U_s$.

In practical, to obtain $B$ and the solution $w_{s+1}$ of \eqref{G2DLDAvorth0},
we first consider to solve the following linear equation
\begin{equation*}
W_s^T G=0.
\end{equation*}
Its solution spans a $(d_1-s)$-dimensional space. We apply the Gram-Schmidt procedure to its solution, and consequently obtain an orthogonal basis $\{b_1,\,b_2,\ldots,b_{d_1-s}\}$ of the linear subspace $U_s^{\bot}$. This implies $B=[b_1,\,b_2,\ldots,b_{d_1-s}]\in\mathbb{R}^{d_1\times (d_1-s)}$ is the solution of $W_s^T G=0$, that is $W_s^T B=0$. After obtaining $B$, let $X_{i}\leftarrow B^TX_{i}$, $V_{i}\leftarrow B^TV_{i}$, $Z_{ij}\leftarrow B^TZ_{ij}$, $i=1,2,\ldots,c$, $j=1,2,\ldots,N_i$. Then we only need to solve $\min\,J_0(w)$ using the updated data. Or equivalently, we solve
\begin{equation}\label{G2DLDAvorth}
\underset{{\widetilde{w}}\in\mathbb{R}^{d_1-s}}{\min}~~
\frac{\sum\limits_{i=1}^{c}\sum\limits_{j=1}^{N_i}\|{\widetilde{w}}^TB^TZ_{ij}\|_p^p+\sigma||{\widetilde{w}}||_p^p}{\sum\limits_{i=1}^{c}N_i\|{\widetilde{w}}^TB^TV_{i}\|_p^p},
\end{equation}
which can be solved by Algorithm 1.

As we can see, the solution $\widetilde{w}^*$ of problem \eqref{G2DLDAvorth} lies in $\mathbb{R}^{d_1-s}$. Now we transform it to an element in $U_s^{\bot}\subseteq\mathbb{R}^{d_1}$.
Since $\mathbb{R}^{d_1-s}$ and $U_s^{\bot}$ are two isomorphic linear spaces, and $B$ is a linear isomorphism between them, then for each $w\in U_s^{\bot}$, $w=\sum\limits_{i=1}^{d_1-s}a_ib_i$ holds, where $a_i\in\mathbb{R}$ is the $i$-th representation coefficient of $w$, and $a_i=<b_i, w>$, where $<b_i, w>$ is the inner product of $b_i$ and $w$. Therefore, $[a_1,\,a_2,\ldots,a_{d_1-s}]^T=[<b_1, w>,<b_2, w>,\ldots,<b_{d_1-s}, w>]^T=[b_1,\,b_2,\ldots,b_{d_1-s}]^Tw=B^Tw$. This implies that multiplying the left side of $w$ by $B^T$ gives the coefficient of the representation by $B$, and we set $w_{s+1}=B\widetilde{w}^*/||B\widetilde{w}^*||_2^2$. \hfill
%The rationality of Algorithm 2 is given in the supplemenary material. In addition, we have the following Proposition.
\begin{center}
\begin{tabular}{lllll}
\toprule
{$\mathbf{Algorithm~2.}$~G2DLDA solving algorithm for multiple}\\
\quad\quad \quad\quad \quad \quad~~{directions}\\
\midrule
\textbf{Input}: The training input data matrix $X \in R^{{d_1}\times d_2}$, \\parameter $\sigma>0$, and the desired discriminant vectors \\ number $r_1$.\\
\textbf{Process}:\\

I. Initialization. Set $W=\varnothing$, and $B=I$;\\

II. \textbf{For} $s=1$ to $r_1$\\
\quad\quad (a) Compute $X_{i}\leftarrow B^TX_{i}$, $V_{i}\leftarrow B^TV_{i}$, and\\
\quad\quad\quad $Z_{ij}\leftarrow B^TZ_{ij}$, $i=1,2,\ldots,c$, $j=1,2,\ldots,N_i$.\\
\quad\quad (b) Apply Algorithm 1 to the updated data in (a),\\
\quad\quad\quad and obtain the optimal solution $\widetilde{w}^*$.\\
%, i.e., \\
%\quad\quad\quad solve problem \eqref{G2DLDAvorth} and obtain $\widetilde{w}^*$;\\
\quad\quad (c) Update $W=[W,w]$, where $w=B\widetilde{w}^*$;\\
\quad\quad (d) Solve the linear equation $W^T G=0$, and\\
\quad\quad\quad update $B$ as the obtained orthogonal solution.\\
\quad\textbf{End}\\
\noindent {\bf Output:} W.\\
\bottomrule
\end{tabular} \label{Algorithm2Rp}\\
\end{center}
\begin{prop} By performing Algorithm 2, the obtained multiple projection vectors are orthogonal to each other.
\end{prop}
\textbf{Proof: }Since $\{b_1,\,b_2,\ldots,b_{d_1-s}\}$ is the orthonormal basis generated by the solution of $W_s^TG=0$, where $W_s=[w_1,\,w_2,\ldots,w_{s}]$, we have $<w_i,w_{s+1}>=w_i^T\cdot w_{s+1}=w_i^TB\widetilde{w}^*=0$, $i=1,2,\ldots, s$, and hence $w_{s+1}\in U_s^{\bot}$.
\section{Experiments}
In this section, experiments are conducted on three contaminated ORL \cite{ORL}, AR \cite{AR}, FERET \cite{FERET} human face databases to evaluate the performance of the proposed G2DLDA compared to 2DPCA \cite{YangZhang04}, 2DPCA-L1 \cite{LiPangYuan10}, 2DLDA \cite{Liuetal93}, and L12DLDA \cite{L12DLDA15,ChenL12DLDA15}. We here write 2DPCA-L1 as L12DPCA for symbol unification.
The learning parameter $\delta$ of L12DLDA is selected optimally from the set $\{0.001, 0.005, 0.01, 0.05, 0.1, 0.5, 1\}$.
For our G2DLDA, we take $p$ from  $\{0.5,1,1.5,2,5\}$, and the parameter $\sigma$ is chosen optimally from the set $\{0.001, 0.01, 0.1, 1, 10\}$. When we examine the influence of the regularization term to our methods, we also consider $\sigma=0$.
In addition, $w^{(0)}$ is initialized as $\textbf{1}/||\textbf{1}||_2$ instead of a random vector in G2DLDA, where $\textbf{1}$ is the vector of all ones, which makes sure that the comparison between our G2DLDA and other methods is not affected by random initialization.
%Since deep network is one of the most powerful state-of-the-art pattern recognition models, we also apply the classical deep model convolutional neural network (CNN) \cite{Krizhevsky12} to the above databases as a baseline.
%The setting of CNN is described as: the net has an input layer of 32x32 neurons followed by a convolution layer with 16 3x3 convolutions; the next hidden layer is a 2x2 max pooling layer with stride [2 2] and padding [0 0 0 0]; its outputs connected to another convolution layer containing 32 maps of 3x3 neurons each, which follows by a 2x2 max pooling, 64 3x3 convolutions and fully connected layer. The output activation function is the softmax function.
%ReLU is used as the active function.
%
%1   ''   Image Input             32x32x1 images with 'zerocenter' normalization
%     2   ''   Convolution             16 3x3 convolutions with stride [1  1] and padding [1  1  1  1]
%     3   ''   ReLU                    ReLU
%     4   ''   Max Pooling             2x2 max pooling with stride [2  2] and padding [0  0  0  0]
%     5   ''   Convolution             32 3x3 convolutions with stride [1  1] and padding [1  1  1  1]
%     6   ''   ReLU                    ReLU
%     7   ''   Max Pooling             2x2 max pooling with stride [2  2] and padding [0  0  0  0]
%     8   ''   Convolution             64 3x3 convolutions with stride [1  1] and padding [1  1  1  1]
%     9   ''   ReLU                    ReLU
%    10   ''   Fully Connected         40 fully connected layer
%    11   ''   Softmax                 softmax
%    12   ''   Classification Output   crossentropyex
All of our experiments are carried out on a PC machine with Intel 3.30 GHz CPU and 4 GB RAM memory under Matlab 2017b platform.
%except for CNN, which is implemented on Linux.
To test the discriminant ability for various methods, we first project test images into a new space obtained by each dimensionality reduction method on training samples; then we apply the nearest neighbor classifier under F-norm metric to classify the test face images. In this section, we give main conclusions. For more detailed descriptions, we refer the readers to the supplemenary materials.

%The model was trained using stochastic gradient descent method, and the learning rate was initialized at 0.01. ReLU was used as the hidden activation function. The net has an input layer of 32x32 neurons followed by a convolution layer with 16 3x3 convolutions. The next hidden layer is a 2x2 max pooling layer with stride [2 2] and padding [0 0 0 0]. Its outputs connected to another convolution layer containing 32 maps of 3x3 neurons each. Its follows by a 2x2 max pooling, 64 3x3 convolutions and 40 fully connected layer. The output activation function is the softmax function.

\subsection{Human face databases}

\begin{figure}[htbp]
\begin{center}{
{
\resizebox*{8cm}{!}
{\includegraphics{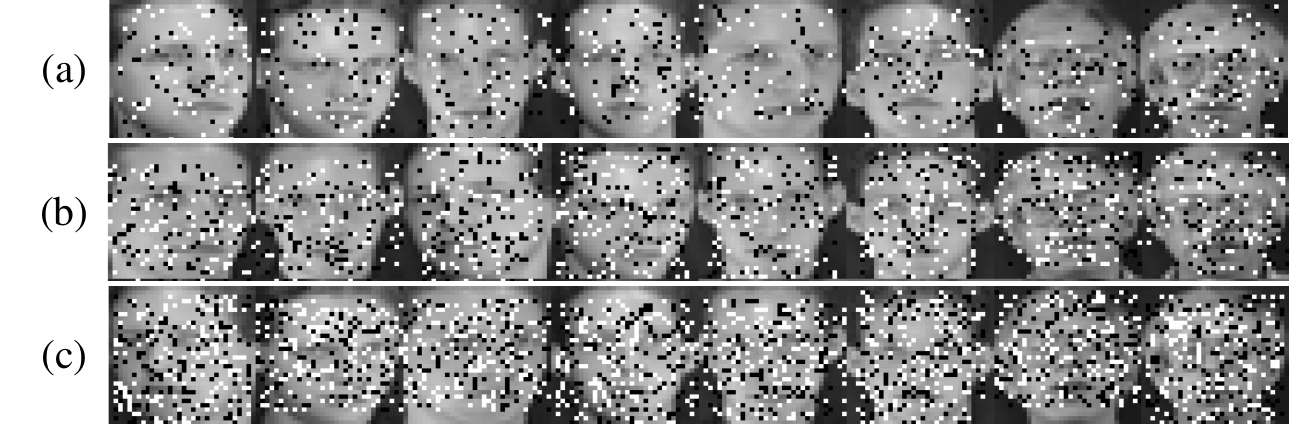}}}\hspace{5pt}
\caption{Sample faces from the ORL database with the random salt and pepper noise to each whole training image with noise densities (a) 0.1, (b) 0.2 and (c) 0.3, respectively.}
\label{ORLsamples}}
\end{center}
\end{figure}

The ORL database contains 400 samples of 40 subjects, where each subject has 10 different images. Some of the images were taken at different times, varying the lighting, facial expressions and facial details. All the images were taken against a dark homogeneous background with the subjects in an upright, frontal position, with pixels 119$\times$92. We here crop and resize each image to the size 32$\times$32. 6 images per subject are randomly selected to formed the training set, and the rest images form the test set. Each training face image is added with the random salt and pepper noise to each whole training image with noise densities 0.1, 0.2 and 0.3, respectively, as shown in Figure \ref{ORLsamples}.

\begin{figure}[htbp]
\begin{center}{
{
\resizebox*{8cm}{!}
{\includegraphics{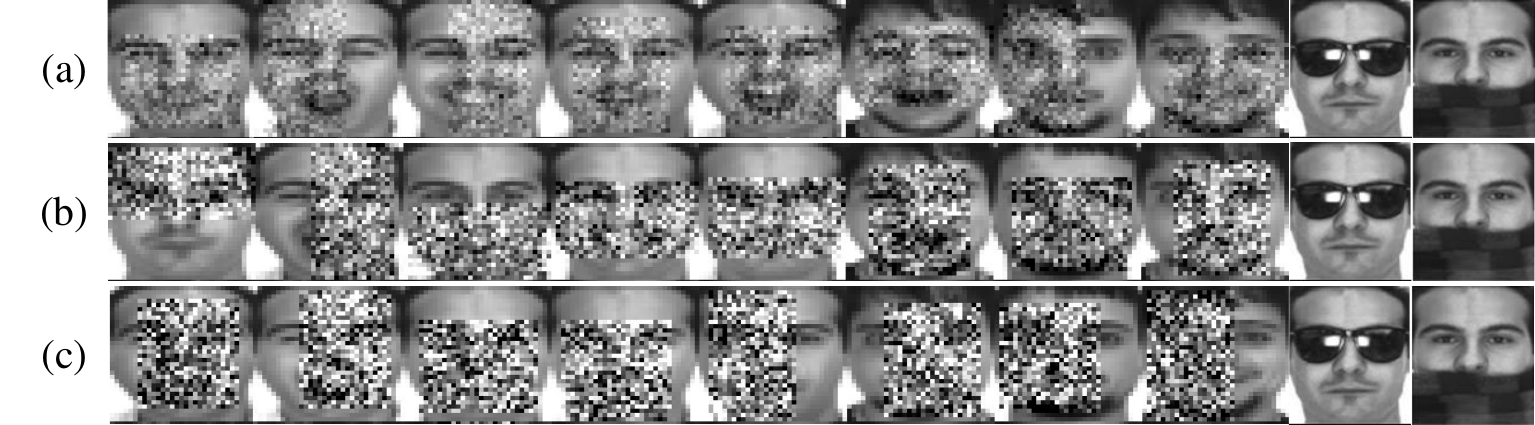}}}\hspace{5pt}
\caption{Sample faces from the AR database with 50\% random rectangular Gaussian noise with mean 0, and variances (a) 0.01, (b) 0.05 and (c) 0.1, respectively. The last two images of each row are the natural occulted faces. }
\label{ARsamples}}
\end{center}
\end{figure}

The AR face database contains 120 subjects with each object containing 26 images of size 50$\times$40. All the images were taken with different facial expressions and illuminations, and some images were occluded with black sunglasses or towels, as shown in Figure \ref{ARsamples}. We here use a subset of the AR database that contains 100 subjects with each subject containing 12 images, and further crop and resize each face image to the size $32\times 32$. Two of the 12 images are occluded with black sunglasses, and two of them are occluded with towels. The rest 8 images of each subject are unoccluded.
For unoccluded faces, we randomly choose 5 images of each subject together with the natural occluded images as the training set, and the rest 3 constitute the test set. The above originally unoccluded training images are artificially polluted with random Gaussian noise of rectangular form with mean 0 and variances 0.01, 0.05 and 0.1, respectively, and the area of noise covers 50\% of each image at random position. The position, the length and width of the noise rectangular are also randomly generated. The sample face images of the polluted training data are shown in Figure \ref{ARsamples}.

\begin{figure}[htbp]
\begin{center}{
{
\resizebox*{8cm}{!}
{\includegraphics{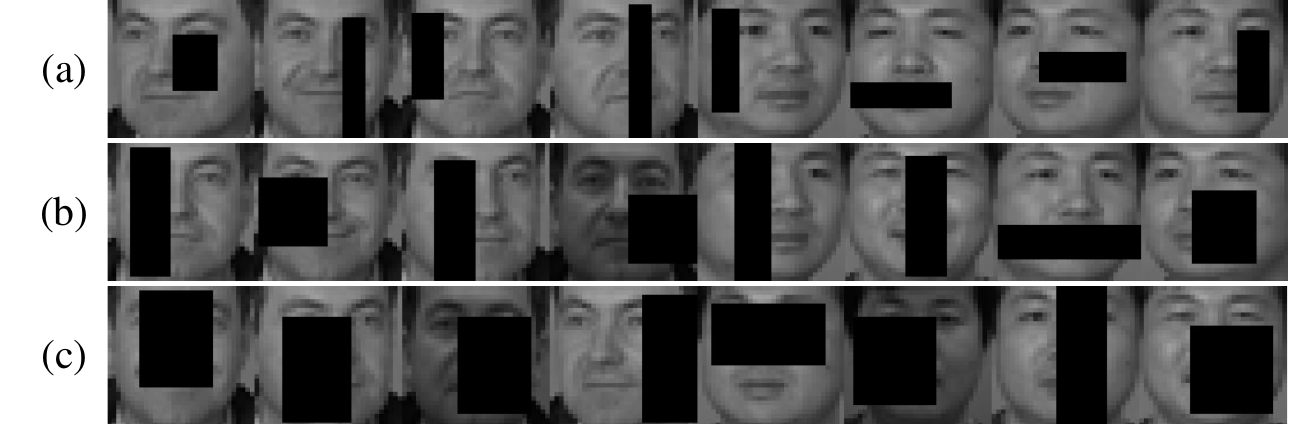}}}\hspace{5pt}
\caption{Sample faces from the FERET database with random rectangular black block covered on (a) 10\%, (b) 20\% and (c) 30\% percentages of each image, respectively.}
\label{FERETsamples}}
\end{center}
\end{figure}

The FERET database contains 1564 sets of images for a total of 14126 images that include 1199 individuals and 365 duplicate sets. Each photography session used the same physical setup, and for some individuals, over two years had elapsed between their first and last sittings, with some subjects being photographed multiple times. We here use a subset that contains 1400 gray scale images of 200 individuals, with each image cropped and resized to 32$\times$32. For each individual there are 7 face images with expression, illumination and age variation.
4 images per subject are randomly selected to formed the training set, and the rest images form the test set. Each training face is added with black block at random position, and the area of noise covers 10\%, 20\% and 30\% respectively, as shown in Figure \ref{FERETsamples}.

\subsection{The influence of $p$}

%\begin{figure*}
%\begin{center}{
%\subfigure[Salt and pepper 0.1]{
%\resizebox*{5.5cm}{!}
%{\includegraphics{p_ORL_s_50_80_10}}}\hspace{5pt}
%\subfigure[Salt and pepper 0.2]{
%\resizebox*{5.5cm}{!}
%{\includegraphics{p_ORL_s_50_80_20.eps}}}\hspace{5pt}
%\subfigure[Salt and pepper 0.3]{
%\resizebox*{5.5cm}{!}
%{\includegraphics{p_ORL_s_50_80_30.eps}}}\hspace{5pt}
%\caption{Accuracies of G2DLDA under different $p$ ($p=0.5,~1,~1.5,~2,~5$) on the contaminated ORL database with random rectangular salt and pepper noise on each whole training data with densities 0.1, 0.2 and 0.3, respectively.}
%\label{pORL}}
%\end{center}
%\end{figure*}

We first investigate the influence of $p$ to our G2DLDA. For each $p$, we record its accuracy when the reduced dimension varies from 1 to 32, and the results are give in Supplemenary Figures S1, S2, S3 (Figure S1, S2, S3 in supplemenary material).
For ORL database, when $p=1$ and $p=1.5$, G2DLDA has the highest accuracy and performs the stablest.
%Moreover, the best accuracies for all the $p$ are almost the same, and a few discriminant directions are usually sufficient for G2DLDA to achieve relatively good behavior.
In particular, when $p=1.5$, its highest accuracy is 94.38\%. When the noise densities are 0.2 and 0.3, the corresponding results suggest the similar performance of G2DLDA. Generally, when $p=1$ and 1.5, G2DLDA behaves the best on the ORL database. For AR and FERET databases, when $p=0.5,~1,~1.5$, G2DLDA performs better overall. In summary, when the value of $p$ is smaller than 2, G2DLDA behaves better.

%\begin{figure*}
%\begin{center}{
%\subfigure[Salt and pepper 0.1]{
%\resizebox*{5.5cm}{!}
%{\includegraphics{sigma_ORL_10.eps}}}\hspace{5pt}
%\subfigure[Salt and pepper 0.2]{
%\resizebox*{5.5cm}{!}
%{\includegraphics{sigma_ORL_20.eps}}}\hspace{5pt}
%\subfigure[Salt and pepper 0.3]{
%\resizebox*{5.5cm}{!}
%{\includegraphics{sigma_ORL_30.eps}}}\hspace{5pt}
%\caption{Accuracies of G2DLDA under different $\sigma$ ($\sigma=0.001,~0.01,~0.1,~1,~10$) on the contaminated ORL database with random rectangular salt and pepper noise on each whole training data with densities 0.1, 0.2 and 0.3, respectively.}
%\label{sigmaORL}}
%\end{center}
%\end{figure*}

\subsection{The influence of the regularization parameter}
Next, we study the effect of the regularization term $\sigma||W||_p^p$ to G2DLDA. For this purpose, we let $\sigma=0, 0.001, 0.01, 0.1, 1, 10$, respectively. The corresponding results are illustrated in Supplemenary Figures S4, S5 and S6.
For ORL database, we take $p=1.5$ for all the noise densities since G2DLDA performs the best for all its noise situations. For all the noise densities, the best accuracies when $\sigma>0$ are better than those of $\sigma=0$, which demonstrates the benifit of adding the regularization term.
When the noise densities are 0.1 and 0.2, $\sigma=0.001$ gives the best results. When the noise density increases to 0.2 or 0.3, $\sigma=0.1$ gives the best performance. The above results show that for the same data set, the optimal regularization parameters for different noise cases are within a certain interval.
The results on the AR and FERET databases confirm the improvement of G2DLDA brought forward by the regularization term, and for the same database with different noise, the optimal $\sigma$ exists in a similar range.

\subsection{The influence of the reduced dimension}

%However, the effect of the regularization term becomes more obvious, which is shown by the fact that the performance of 2DLDA when $\sigma>0$ is much better than the case $\sigma=0$.

\begin{figure*}[htbp]
\begin{center}{
\subfigure[2DPCA]{
\resizebox*{5.5cm}{!}
{\includegraphics{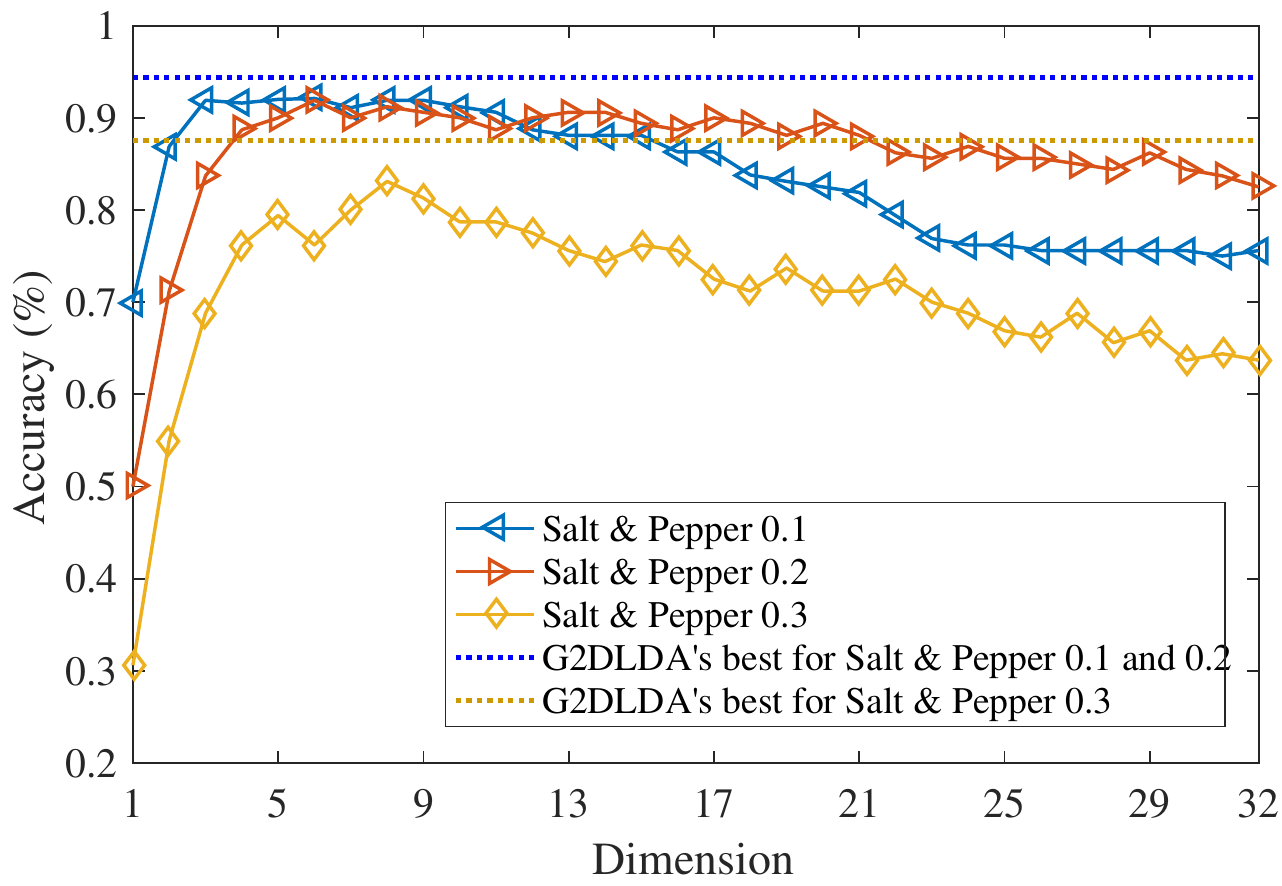}}}\hspace{5pt}
\subfigure[L12DPCA]{
\resizebox*{5.5cm}{!}
{\includegraphics{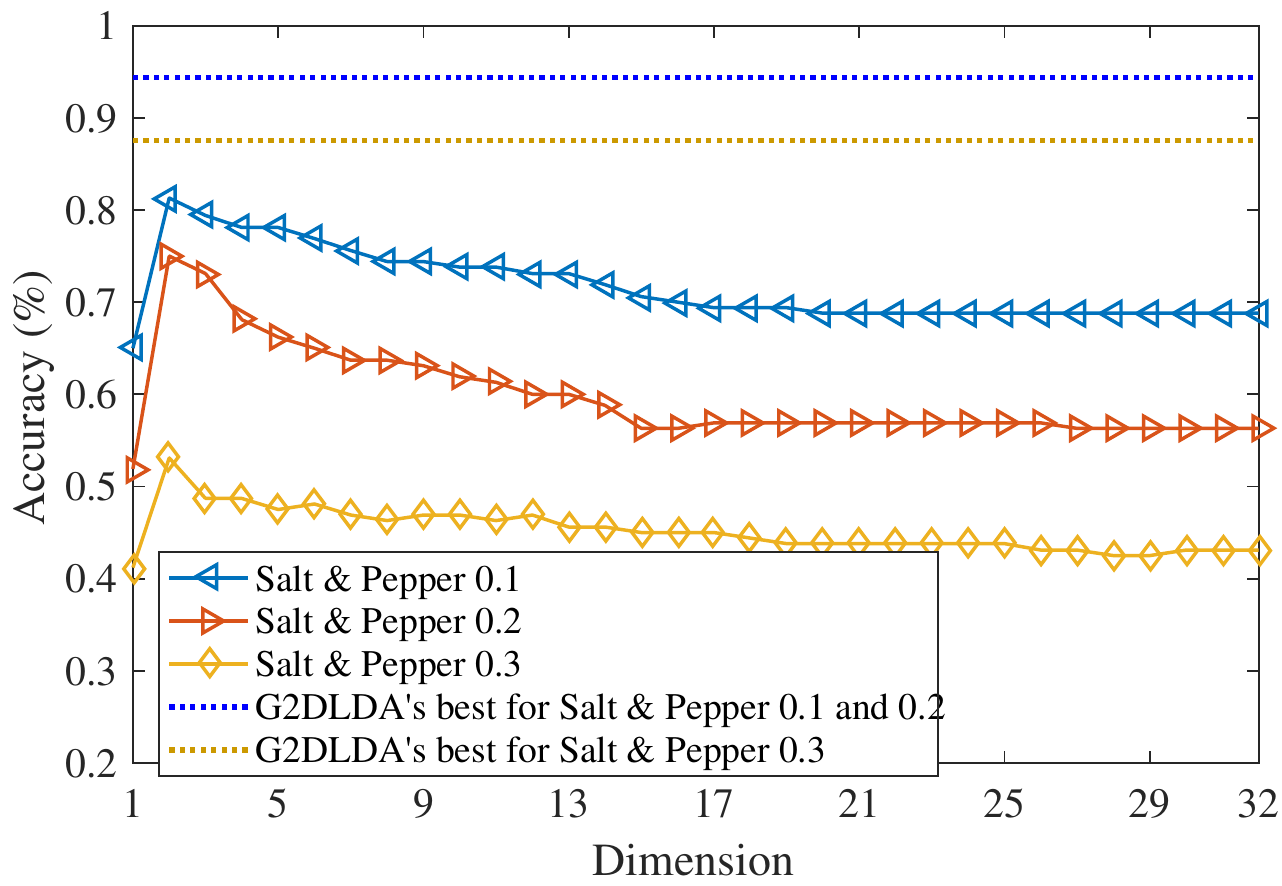}}}\hspace{5pt}
\subfigure[2DLDA]{
\resizebox*{5.5cm}{!}
{\includegraphics{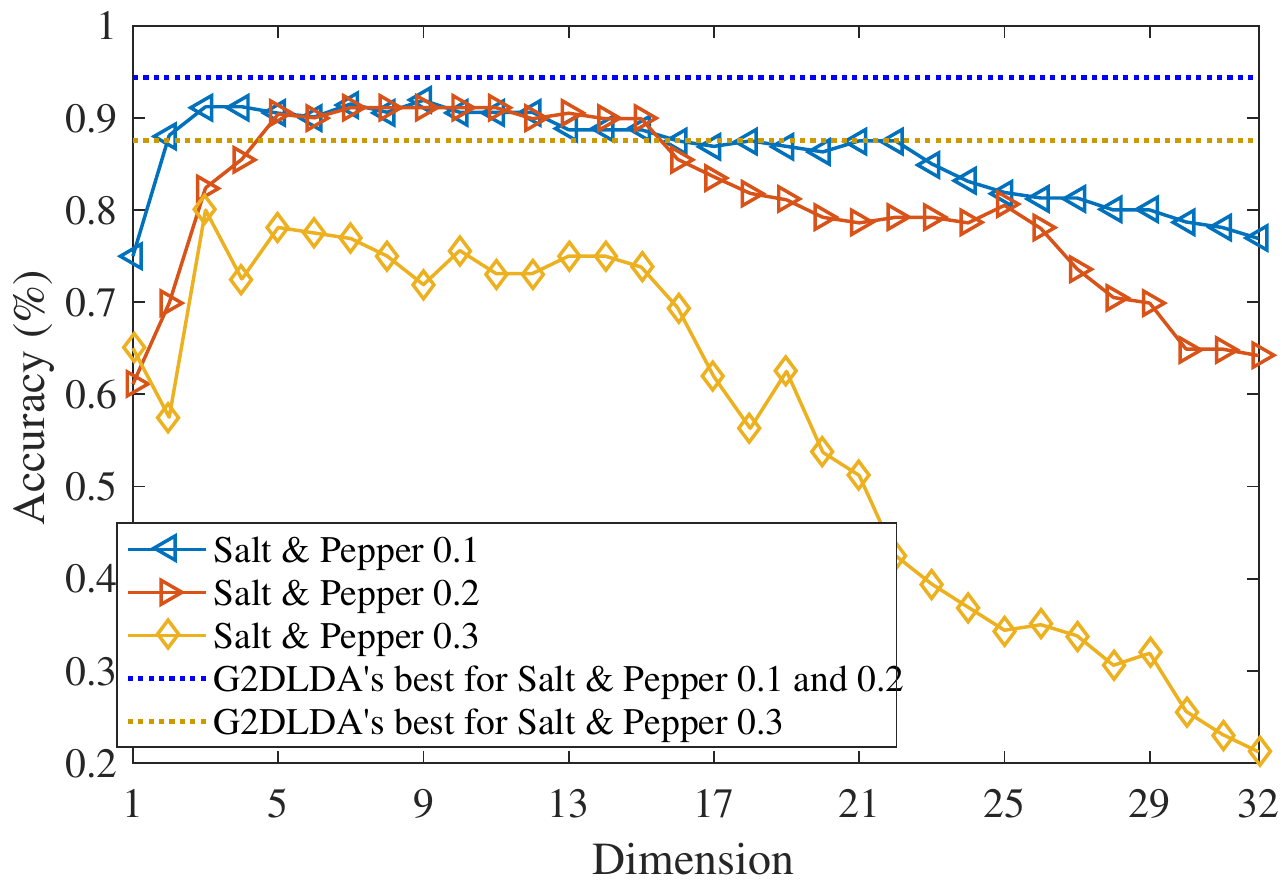}}}\hspace{5pt}
\subfigure[L12DLDA]{
\resizebox*{5.5cm}{!}
{\includegraphics{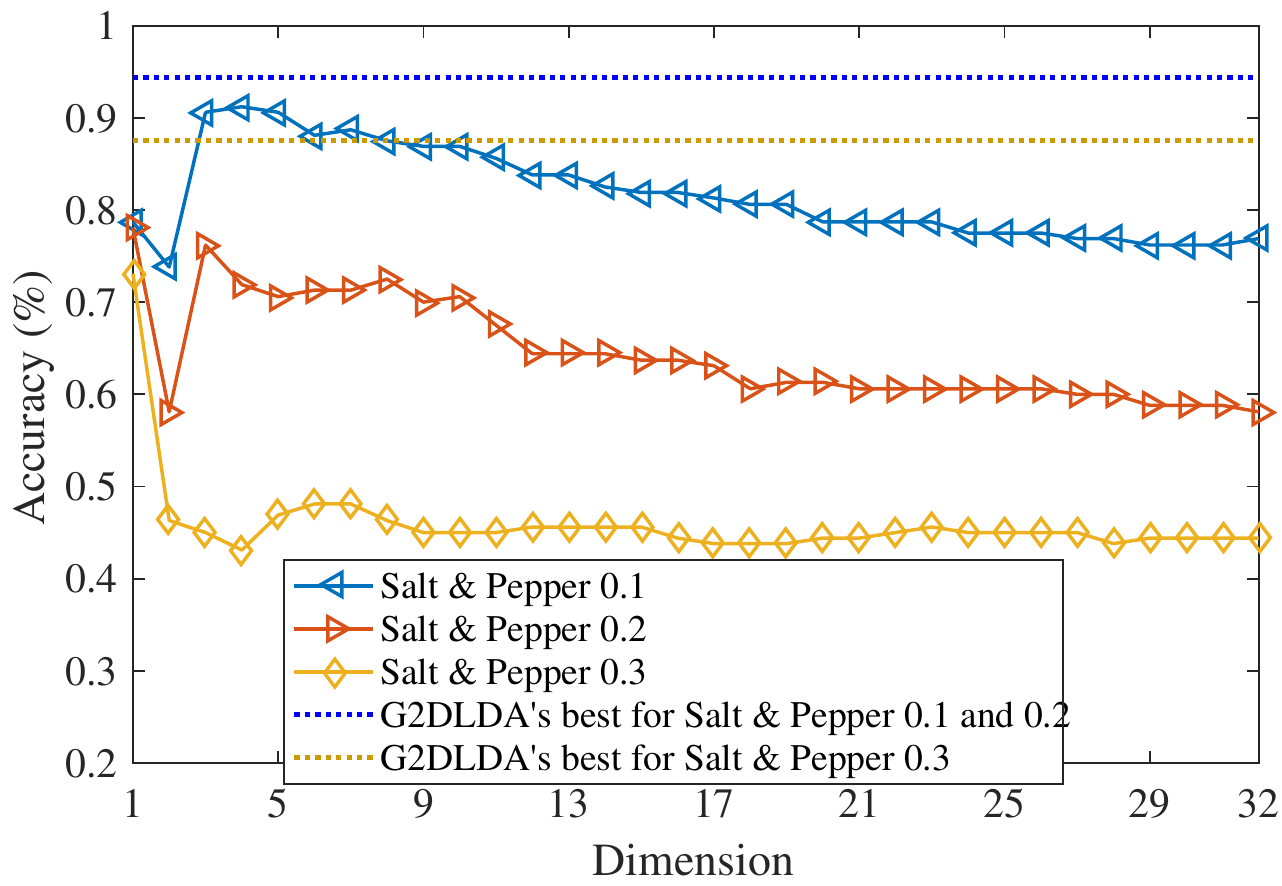}}}\hspace{5pt}
\subfigure[G2DLDA]{
\resizebox*{5.5cm}{!}
{\includegraphics{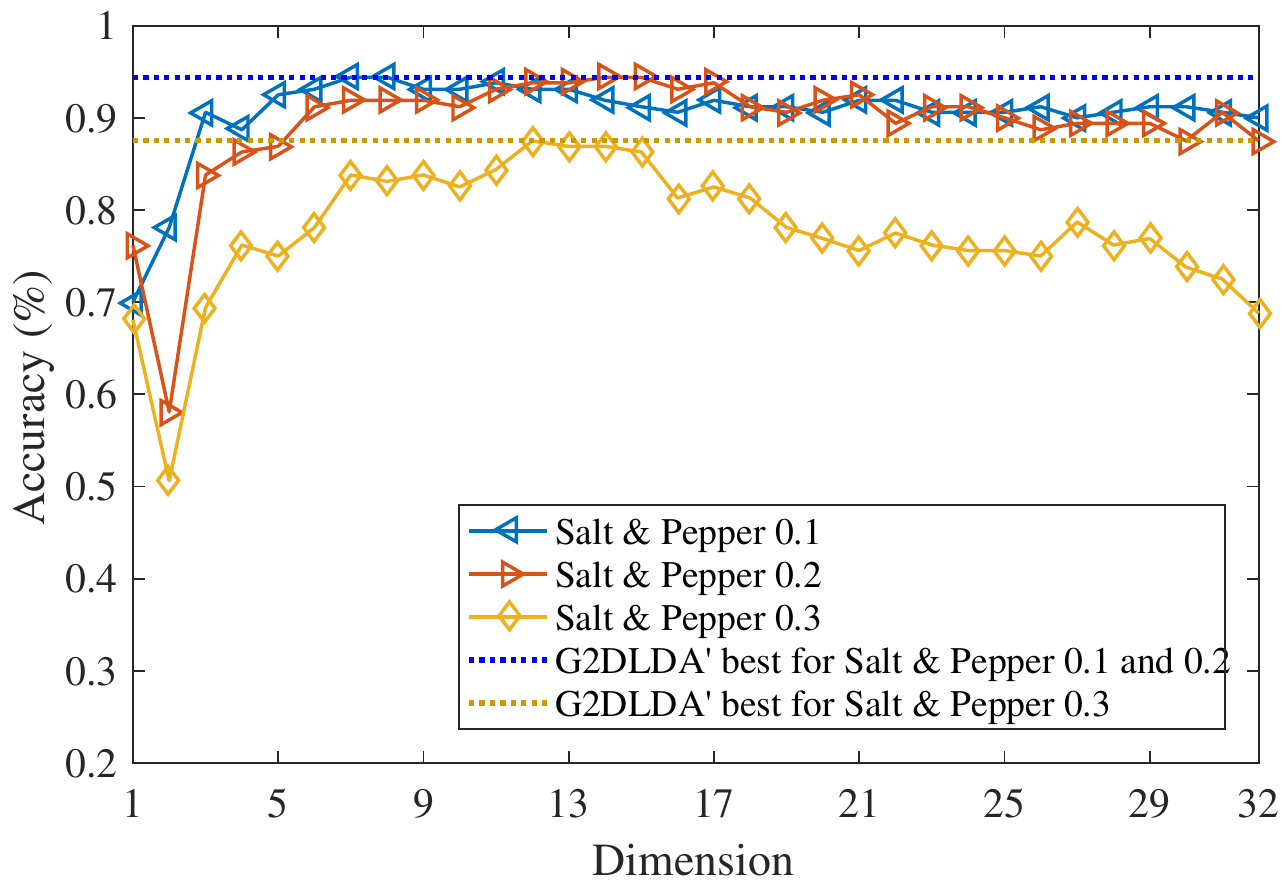}}}\hspace{5pt}
\caption{Accuracies of 2DPCA, L12DPCA, 2DLDA, L12DLDA and G2DLDA under different reduced dimensions on the contaminated ORL database with random rectangular salt and pepper noise on each whole training data with densities 0.1, 0.2 and 0.3, respectively.}
\label{ORLac}}
\end{center}
\end{figure*}

\begin{figure*}[htbp]
\begin{center}{
\subfigure[2DPCA]{
\resizebox*{5.5cm}{!}
{\includegraphics{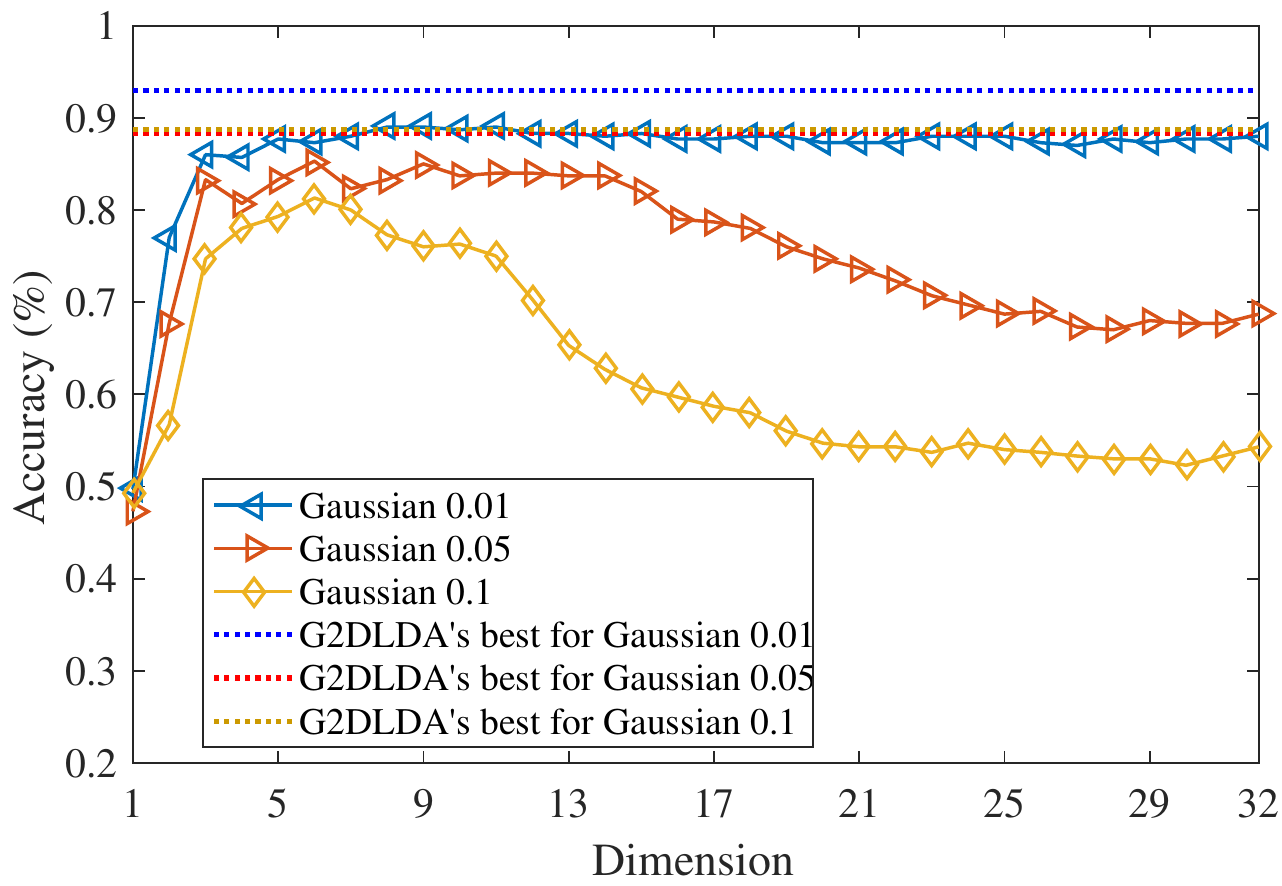}}}\hspace{5pt}
\subfigure[L12DPCA]{
\resizebox*{5.5cm}{!}
{\includegraphics{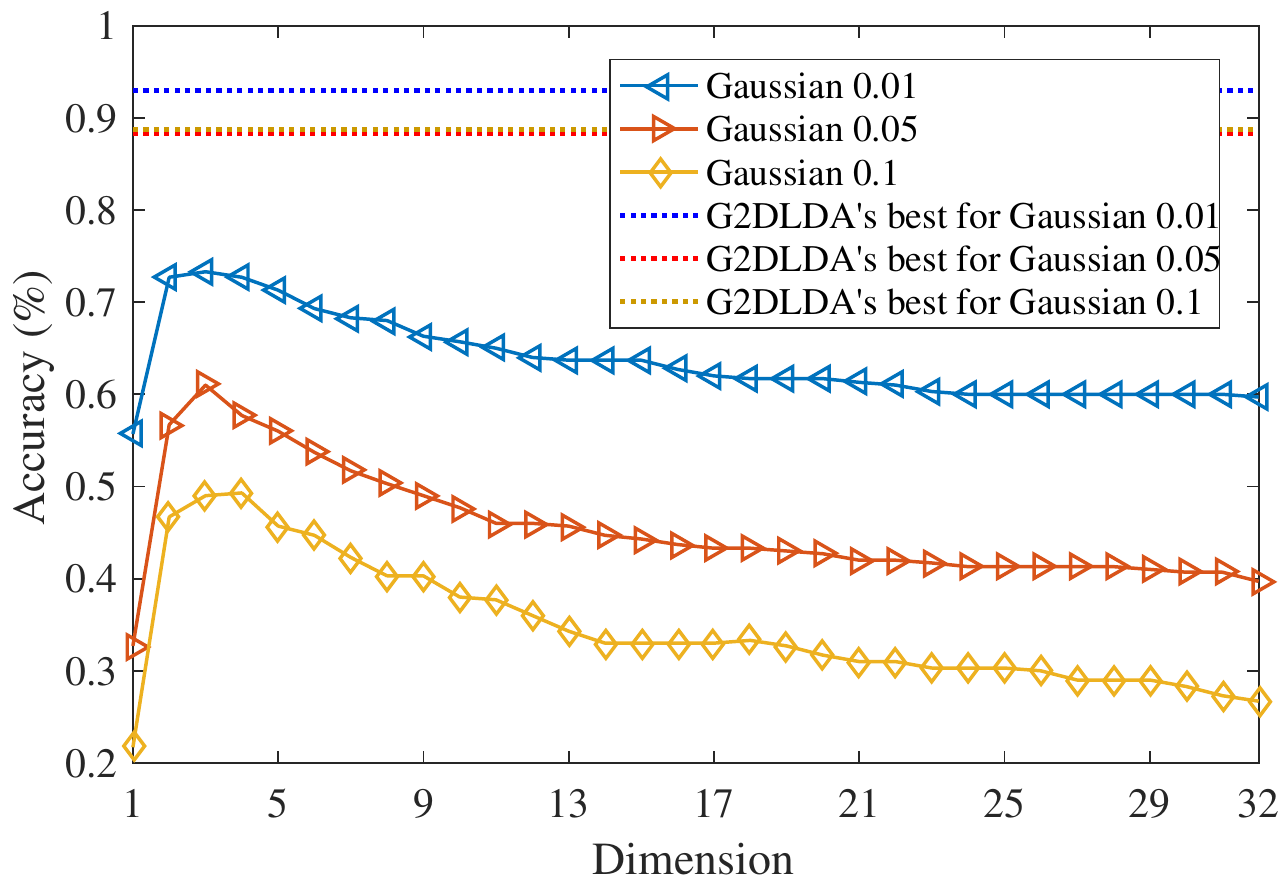}}}\hspace{5pt}
\subfigure[2DLDA]{
\resizebox*{5.5cm}{!}
{\includegraphics{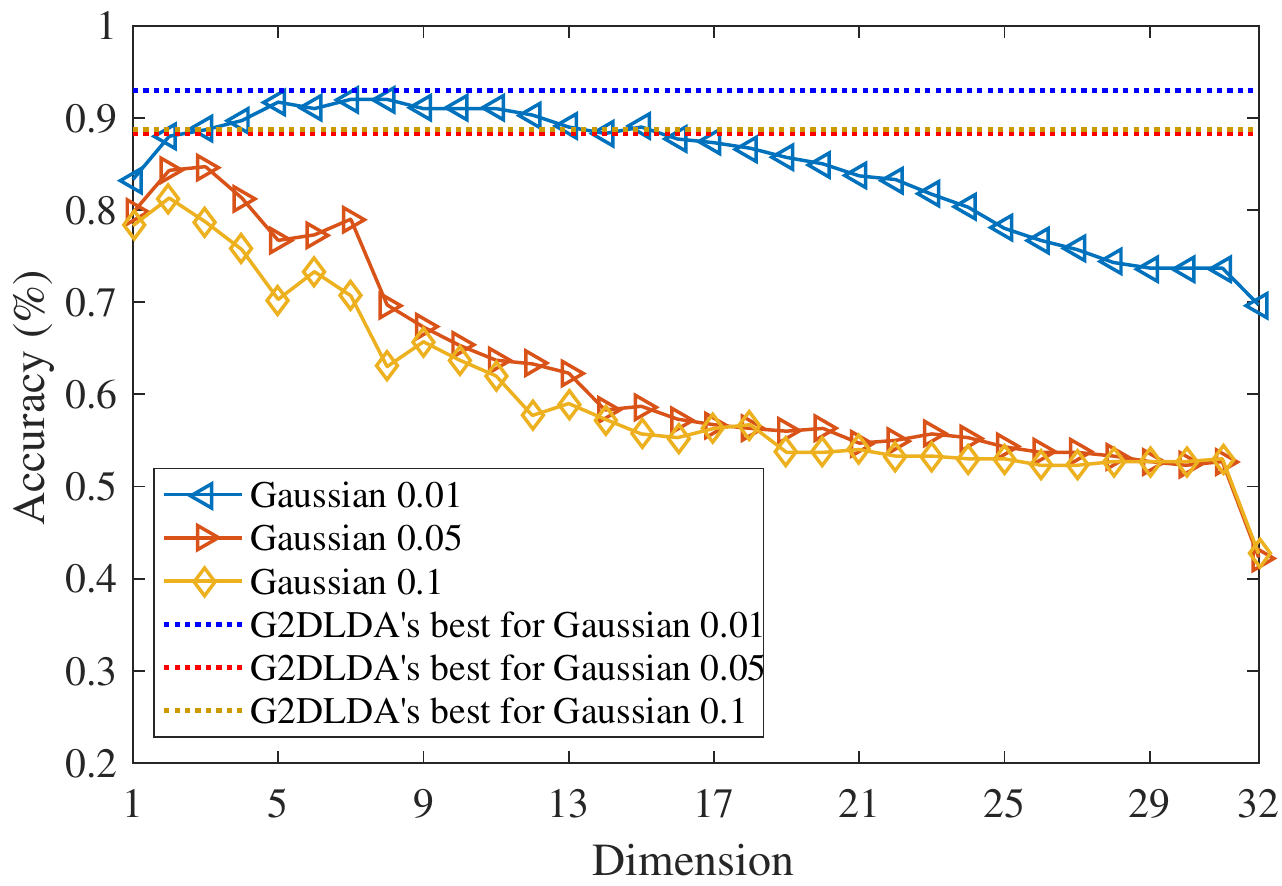}}}\hspace{5pt}
\subfigure[L12DLDA]{
\resizebox*{5.5cm}{!}
{\includegraphics{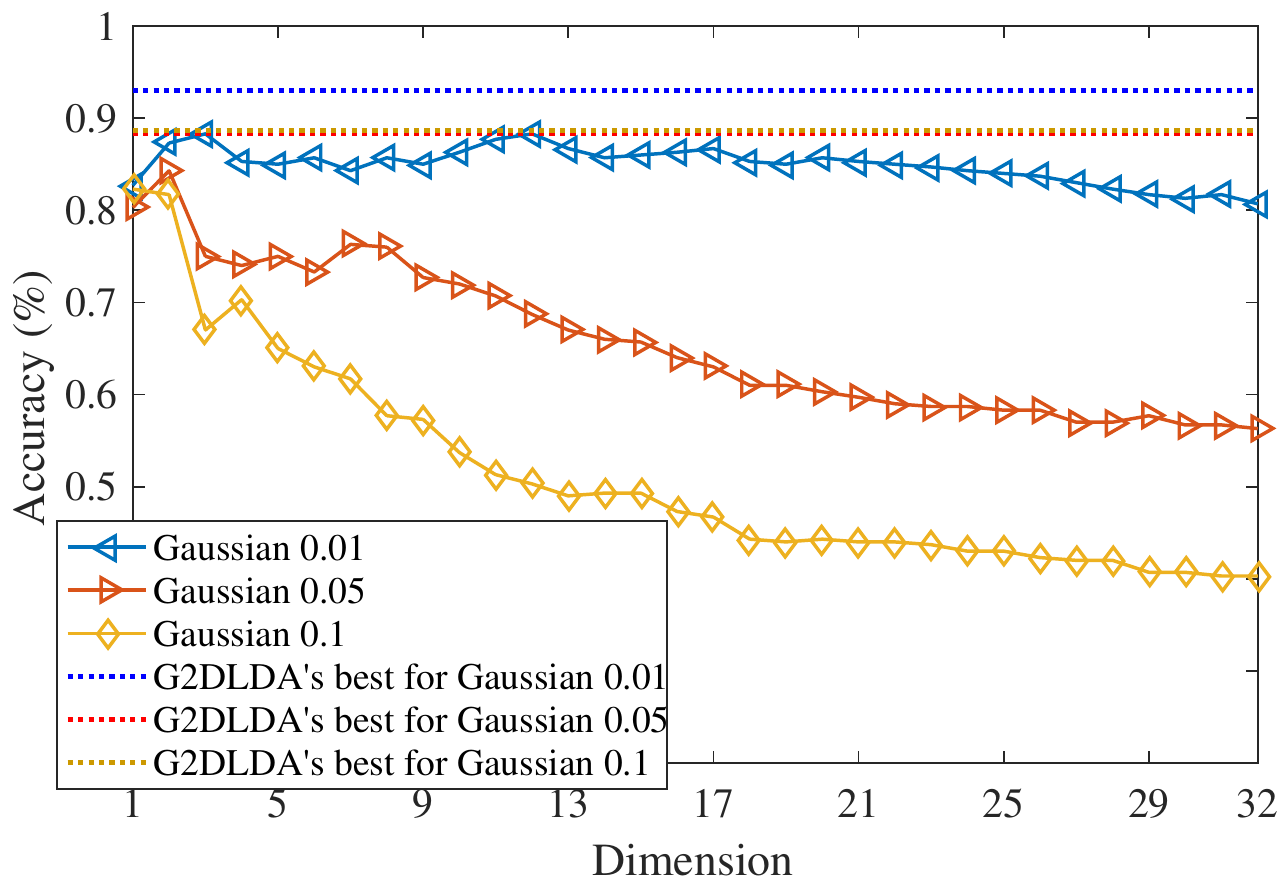}}}\hspace{5pt}
\subfigure[G2DLDA]{
\resizebox*{5.5cm}{!}
{\includegraphics{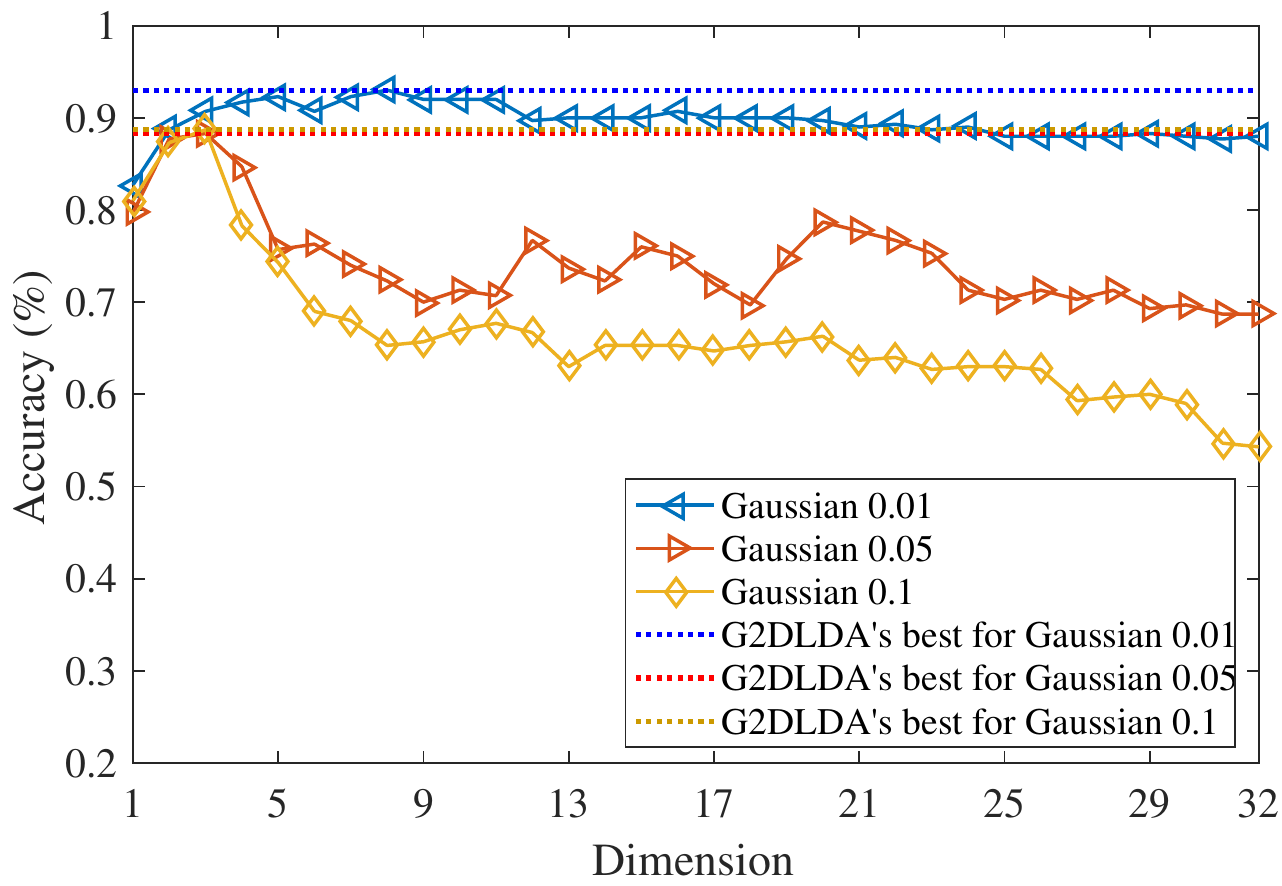}}}\hspace{5pt}
\caption{Accuracies of 2DPCA, L12DPCA, 2DLDA, L12DLDA and G2DLDA under different reduced dimensions on the contaminated AR database with 50\% random rectangular Gaussian noise on the training data with variances 0.01, 0.05 and 0.1, respectively.}
\label{ARac}}
\end{center}
\end{figure*}

\begin{figure*}[htbp]
\begin{center}{
\subfigure[2DPCA]{
\resizebox*{5.5cm}{!}
{\includegraphics{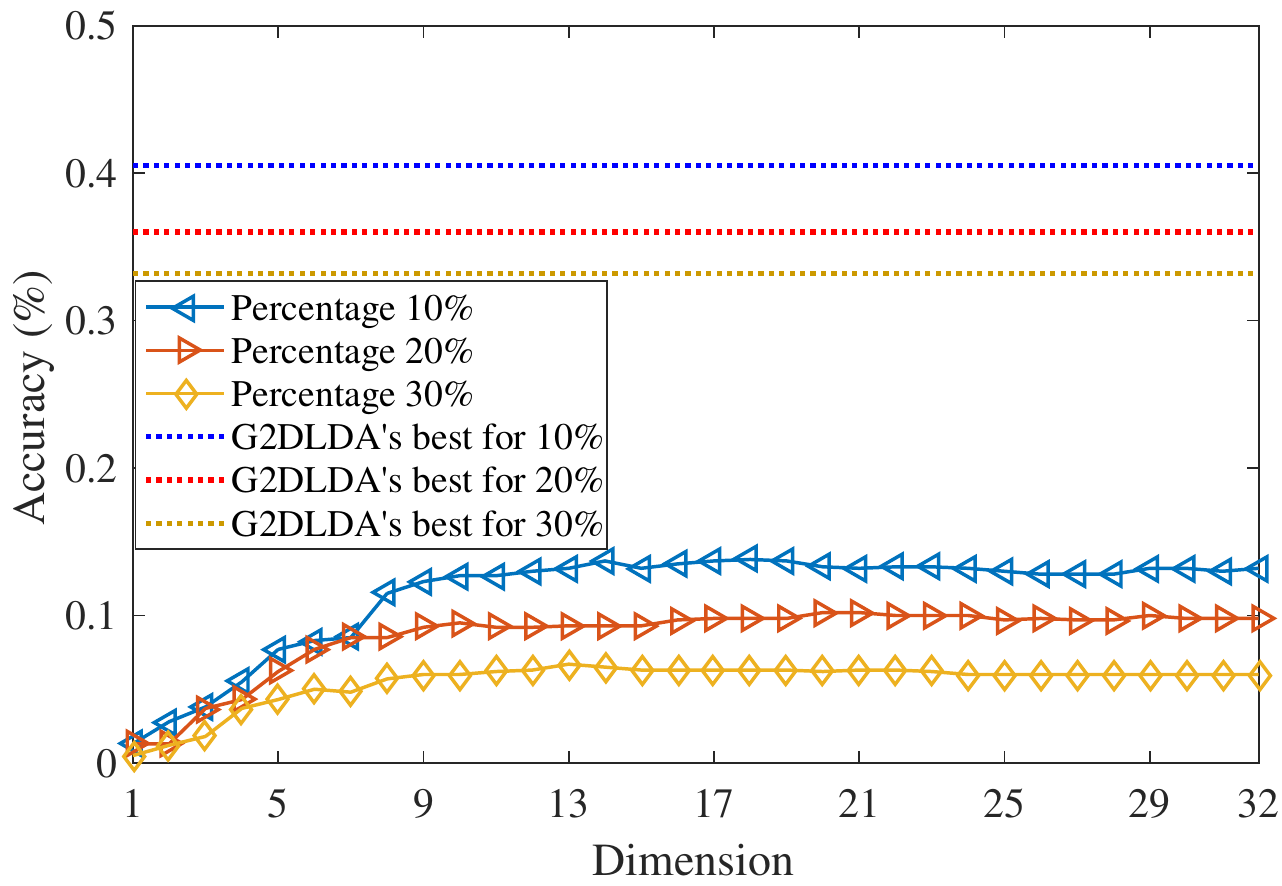}}}\hspace{5pt}
\subfigure[L12DPCA]{
\resizebox*{5.5cm}{!}
{\includegraphics{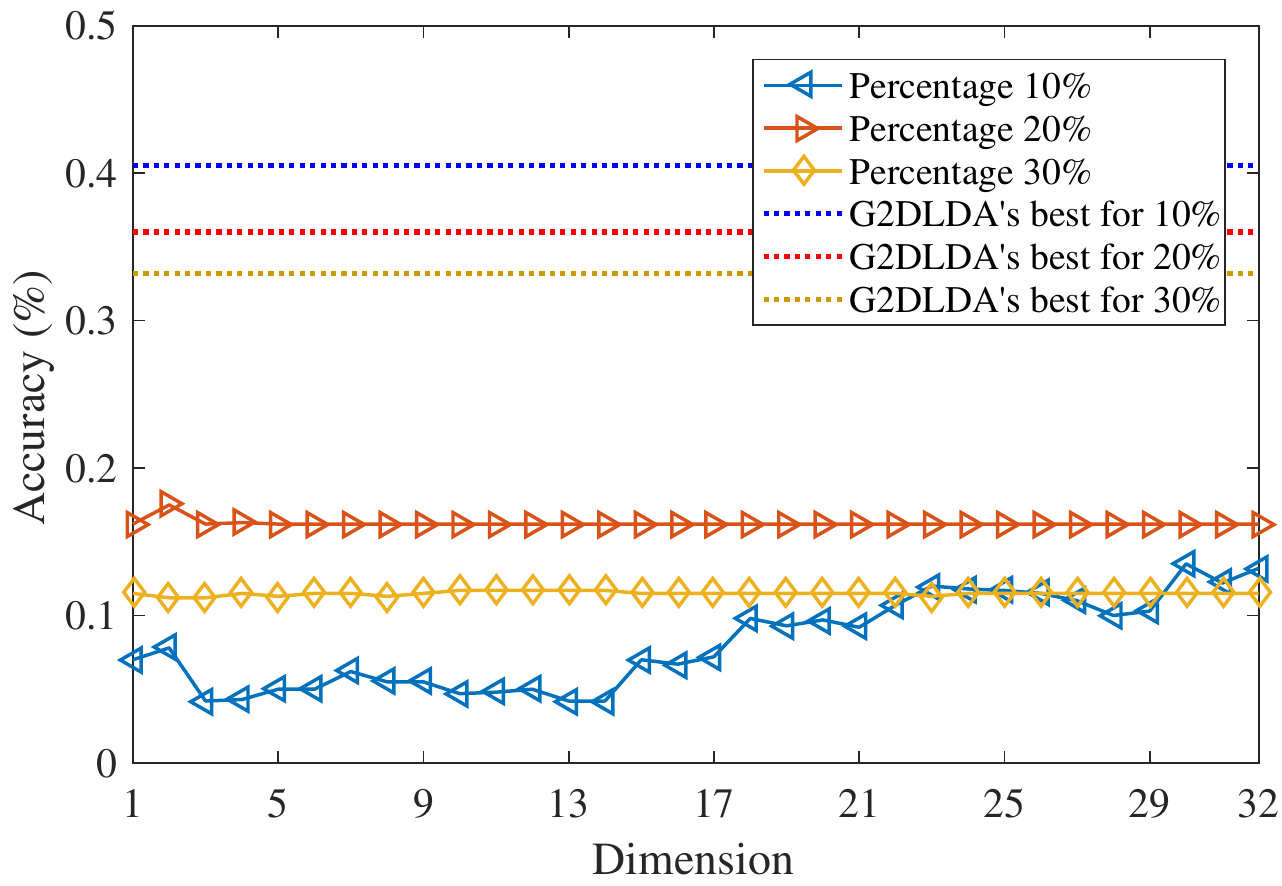}}}\hspace{5pt}
\subfigure[2DLDA]{
\resizebox*{5.5cm}{!}
{\includegraphics{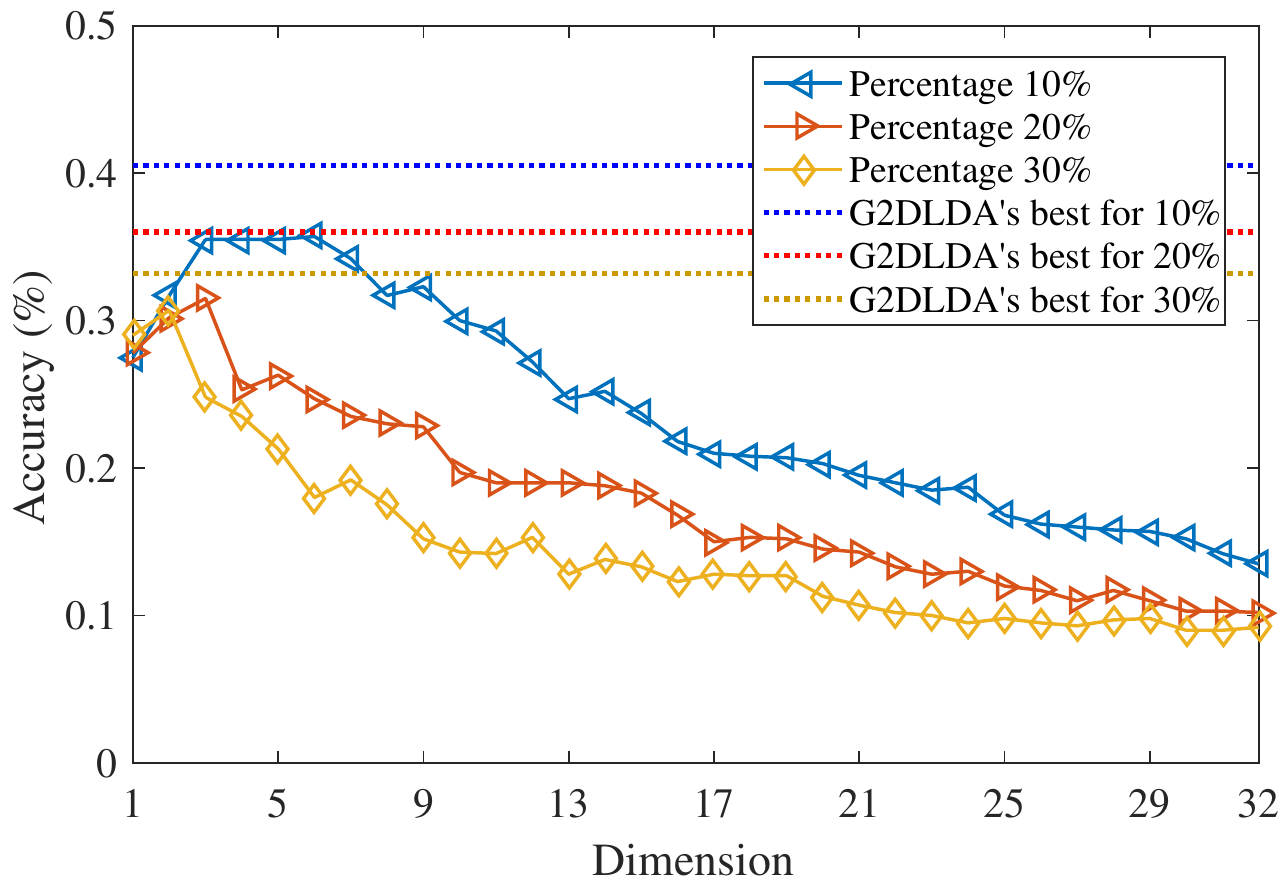}}}\hspace{5pt}
\subfigure[L12DLDA]{
\resizebox*{5.5cm}{!}
{\includegraphics{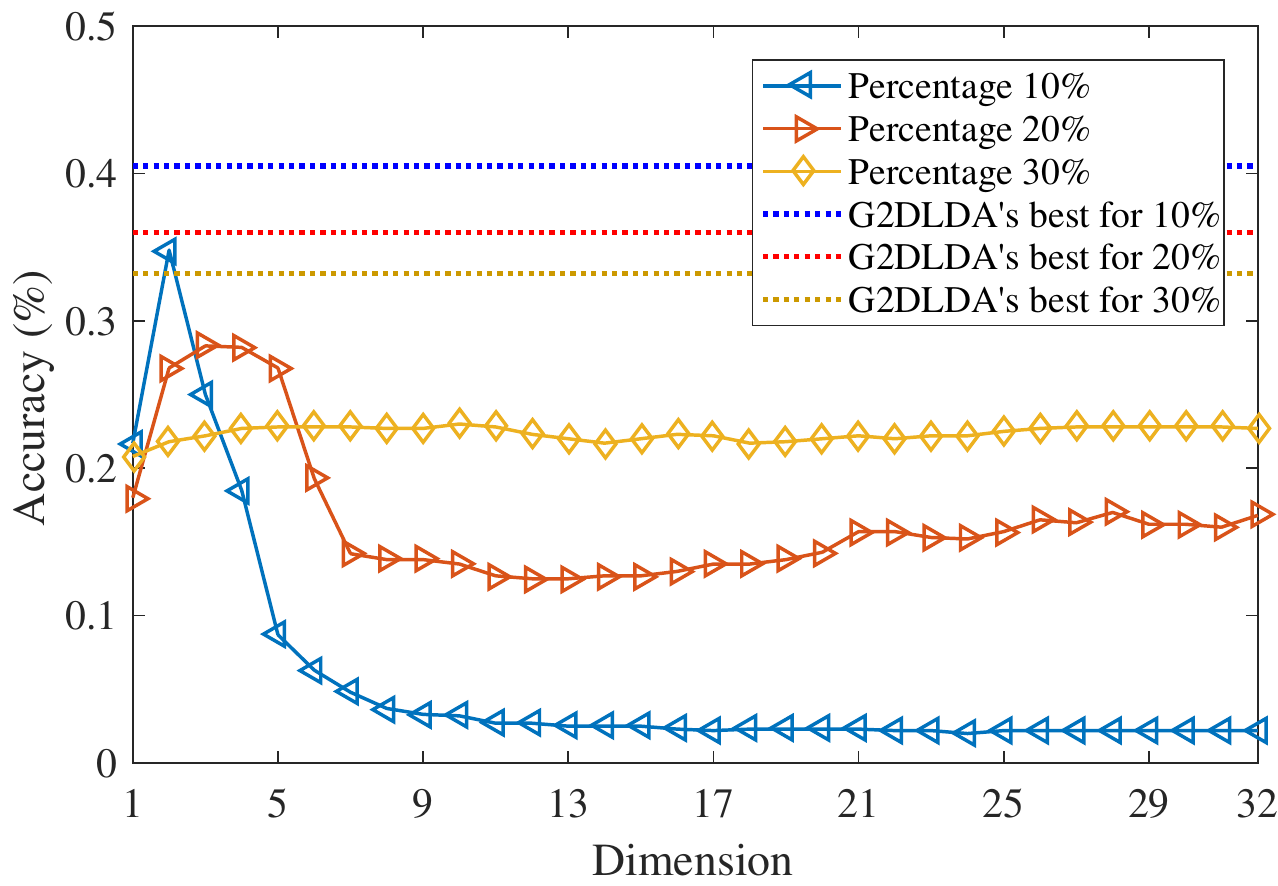}}}\hspace{5pt}
\subfigure[G2DLDA]{
\resizebox*{5.5cm}{!}
{\includegraphics{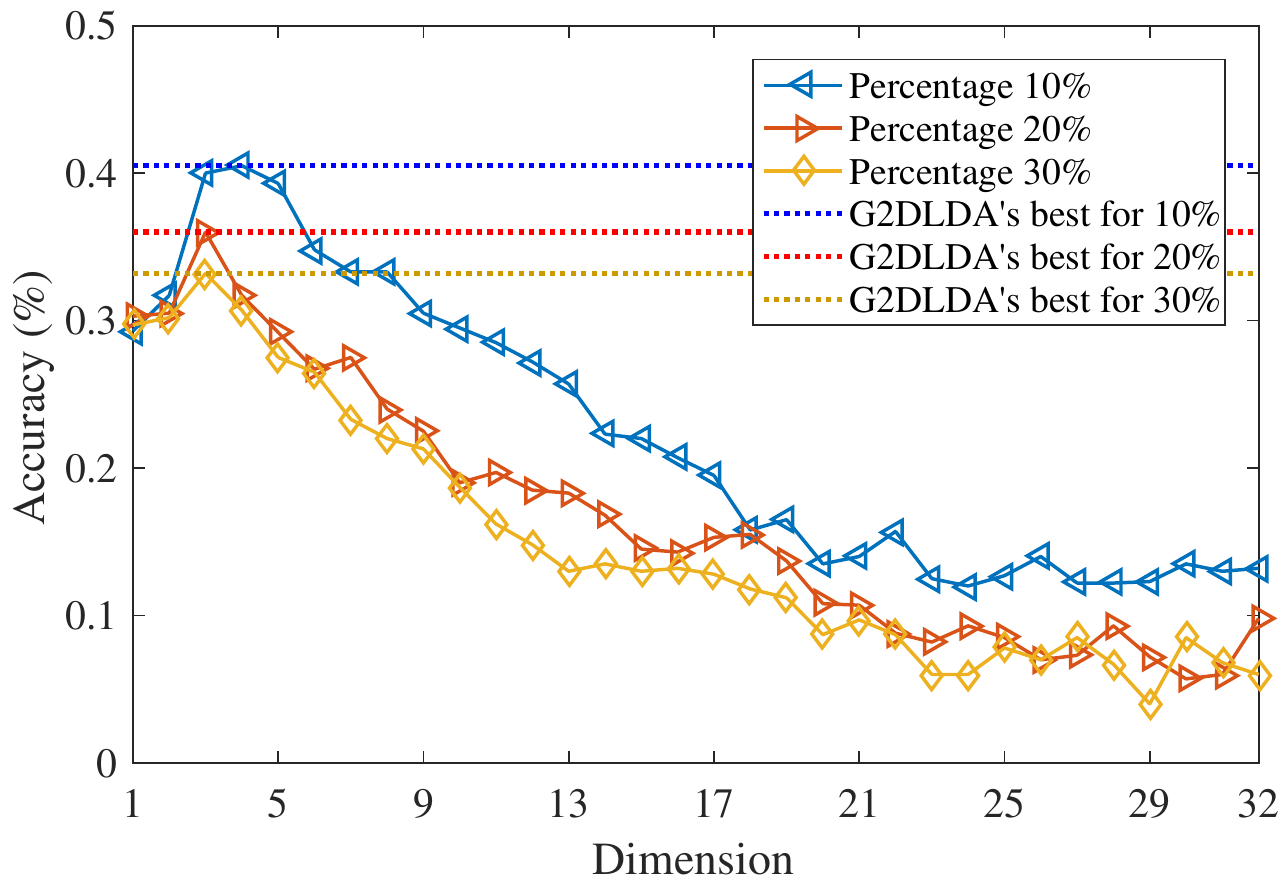}}}\hspace{5pt}
\caption{Accuracies of G2DLDA ($p=0.5,~1,~1.5,~2,~5$) on the contaminated FERET database with rectangular black block covered on 10\%, 20\% and 30\% percentages of each image, respectively.}
\label{FERETac}}
\end{center}
\end{figure*}

We finally examine the behavior of each method when the reduced dimension varies. The corresponding $p$ for each data database is taken as above.
For the ORL database, the result is given in Figure \ref{ORLac}. It is obvious that for all the methods, their overall behaviors become poor as the noise density increases. And as the increasing of the reduced dimension, they all grow fast to the highest accuracy and then descend. This phenomenon shows that within the first a few dimensions, the usable discriminant information increases as the dimension grows. However, as the dimension becomes larger, useless disturbance information may also be included since the image data is polluted.
However, its influence to our G2DLDA is smaller than those of the other methods.
Moreover, the highest accuracies of G2DLDA for all the density levels are better than the other methods.
To see the results more clearer, we list the highest accuracies of all the
methods in Table \ref{TableORL}. From the table, we see for all the three densities of noise, when $p=1$ and 1.5, G2DLDA outperforms the other methods, and its best accuracy is at least 3\% higher. This also shows the necessity to choose a proper $p$.
%It should be also noted that G2DLDA performs better than CNN on this database, which shows CNN cannot capture features well when facing salt and pepper noise. This also supports the robustness of G2DLDA.
We also conduct the similar experiments on the AR database and the FERET database, and their results are given in Figures \ref{ARac}, \ref{FERETac} and Tables \ref{TableAR} and \ref{TableFERET}. The corresponding results again demonstrate that our G2DLDA has the most robust performance.

We summarize all the experimental results in the Supplemenary material.

\begin{table}[htbp]
\begin{center}
%\centering
\caption{Comparison of different methods in terms of the best accuracies(\%) on the contaminated ORL database with 50\% rectangular random salt and pepper noise on each whole face image, with densities 0.1, 0.2 and 0.3, respectively. The optimal dimension is shown next to its corresponding accuracy.}
\resizebox{3.5in}{!}
{
\begin{tabular}{lcccccc}
\toprule
&\vline&\multicolumn{5}{c}{Noise density}\\
\cline{3-7}
Method&\vline & 0.1&\vline& 0.2&\vline&0.3\\
\cline{3-7}
&\vline & Acc (Dim)&\vline& Acc (Dim)&\vline&Acc (Dim)\\
\toprule
%CNN&\vline &83.13 &\vline&81.25 &\vline& 76.25\\
2DPCA&\vline &91.88 (6) &\vline& 91.88 (6)&\vline& 83.13 (8)\\
L12DPCA&\vline &81.25 (2)&\vline& 75.00 (2)&\vline& 53.13 (2)\\
2DLDA&\vline  &91.88 (9)&\vline& 91.25 (7)&\vline& 80.00 (3)\\
L12DLDA&\vline  &91.25 (4)&\vline& 78.13 (1)&\vline&73.13 (1)\\
G2DLDA (p=0.5)&\vline & 91.88 (26) &\vline&88.75 (26) &\vline&73.75(1) \\
G2DLDA (p=1)&\vline & 93.75 (11) &\vline&93.75 (19) &\vline&85.00 (17) \\
G2DLDA (p=1.5)&\vline &\textbf{94.38} (7) &\vline&\textbf{94.40} (14) &\vline&\textbf{87.50} (12) \\
G2DLDA (p=2)&\vline & 92.50 (15) &\vline&91.88 (15) &\vline&77.50 (21) \\
G2DLDA (p=5)&\vline & 91.25 (10) &\vline&92.50 (9) &\vline&76.25 (19) \\
\bottomrule
\end{tabular}
}
\label{TableORL}
\end{center}
\end{table}

\begin{table}[htbp]
\begin{center}
%\centering
\caption{Comparison of different methods in terms of the best accuracies(\%) on the contaminated AR database with 50\% rectangular random Gaussian noise with mean 0, and standard variances 0.01, 0.05 and 0.1, respectively. The optimal dimension is shown next to its corresponding accuracy.}
\resizebox{3.5in}{!}
{
\begin{tabular}{lcccccc}
\toprule
&\vline&\multicolumn{5}{c}{Noise level}\\
\cline{3-7}
Method&\vline & 0.01&\vline& 0.05&\vline&0.1\\
\cline{3-7}
&\vline & Acc (Dim)&\vline& Acc (Dim)&\vline&Acc (Dim)\\
\toprule
%CNN&\vline &82.50 &\vline&79.88 &\vline& 73.75\\
2DPCA&\vline &89.00 (8) &\vline&85.33 (6)&\vline&81.33 (6) \\
L12DPCA&\vline &73.33 (3)&\vline&61.00 (3)&\vline&49.33 (4) \\
2DLDA&\vline  &92.00 (7)&\vline&84.67 (3)&\vline&81.33 (2) \\
L12DLDA&\vline  &88.33 (3)&\vline&84.33 (2)&\vline&82.33 (1) \\
G2DLDA (p=0.5)&\vline & 92.00 (4) &\vline&81.67 (4) &\vline&79.00 (2) \\
G2DLDA (p=1)&\vline & 92.00 (4) &\vline&\textbf{88.33 }(3) &\vline&\textbf{88.67} (3) \\
G2DLDA (p=1.5)&\vline & \textbf{93.00} (8) &\vline&83.67 (2) &\vline&85.67 (3) \\
G2DLDA (p=2)&\vline &88.33 (29) &\vline&73.67 (7) &\vline&70.00 (8) \\
G2DLDA (p=5)&\vline &88.67 (26) &\vline&71.67 (24) &\vline&54.67 (10) \\
%&\vline &p=1&\vline& &\vline& \\
\bottomrule
\end{tabular}
}
\label{TableAR}
\end{center}
\end{table}

\begin{table}[htbp]
\begin{center}
%\centering
\caption{Comparison of different methods in terms of the best accuracies(\%) on the contaminated FERET database with rectangular black block covered on 10\%, 20\% and 30\% percentages of each image, respectively. The optimal dimension is shown next to its corresponding accuracy.}
\resizebox{3.5in}{!}
{
\begin{tabular}{lcccccc}
\toprule
&\vline&\multicolumn{5}{c}{Noise percentage}\\
\cline{3-7}
Method&\vline & 10\% &\vline& 20\% &\vline&30\% \\
\cline{3-7}
&\vline & Acc (Dim)&\vline& Acc (Dim)&\vline&Acc (Dim)\\
\toprule
%CNN&\vline &53.25 &\vline&40.63 &\vline&36.39 \\
2DPCA&\vline &13.83 (18) &\vline&10.17 (20) &\vline& 6.67 (13)\\
L12DPCA&\vline &13.50 (30) &\vline&17.50 (2) &\vline& 11.67 (10)\\
2DLDA&\vline  &35.67 (6) &\vline&31.50 (3) &\vline& 30.67 (2)\\
L12DLDA&\vline  &34.83 (2) &\vline&28.33 (3) &\vline& 23.00 (10)\\
G2DLDA (p=0.5)&\vline &13.67 (30) &\vline&30.83 (1) &\vline&29.00 (1) \\
G2DLDA (p=1)&\vline & \textbf{40.50} (4) &\vline&\textbf{36.00} (3)  &\vline&\textbf{33.17} (3) \\
G2DLDA (p=1.5)&\vline &37.67 (3) &\vline&32.00 (3) &\vline&32.17 (3) \\
G2DLDA (p=2)&\vline & 24.67 (23) &\vline&17.17 (26) &\vline&14.50 (15) \\
G2DLDA (p=5)&\vline &29.50 (11) &\vline&18.00 (7) &\vline&12.67 (9) \\
\bottomrule
\end{tabular}
}
\label{TableFERET}
\end{center}
\end{table}

\section{Conclusion}

In this paper, a novel generalized Lp-norm ($p>0$) two-dimensional linear discriminant analysis with regularization (G2DLDA) has been proposed.
G2DLDA is a generalized framework, and can be solved by a simple iterative algorithm. Comparing to 2DLDA and L12DLDA, G2DLDA can avoid the singularity problem, and is more robust to outliers. Moreover, the regularization term also improves its generalization performance. Experimental results confirmed its effectiveness. The corresponding G2DLDA Matlab code and slide can be downloaded from
http://www.optimal-group.org/Resources/Code/G2DLDA.html.

It should be noted that, the regulation term $||W||_p^p$ can be replaced by any s-norm term $||W||_s^s$ for $s>0$, rather than the same Lp-norm as in the between-class scatter and the within-class scatter. In the future, we expect to extend G2DLDA to its nonlinear case, and study the convergent result of the algorithm for $0<p<1$ and $p>2$. The bilateral G2DLDA is also interesting.

\section*{Acknowledgment}

%The authors are grateful to the referee for his/her thorough reading the
%manuscript and for valuable comments and suggestions which have improved
%the presentation of this work.
This work is supported by the National Natural Science Foundation of
China (No.61703370, No.61866010, No.11871183 and No.61603338), the Natural Science Foundation of Zhejiang Province (No.LQ17F030003 and No.LY18G010018), the Natural Science Foundation of Hainan Province  (No.118QN181), and the Foundation of China Scholarship Council (No. 201708330179).
%, and the Scientific Research Fund of Zhejiang Provincial Education Department (No.Y201534889).

\ifCLASSOPTIONcaptionsoff
  \newpage
\fi

\end{document}